%% file: main.tex
\documentclass[letterpaper, 10 pt, journal, twoside]{IEEEtran} 

\pdfoutput=1
\usepackage[T1]{fontenc}
\usepackage[english]{babel}
\usepackage[pdftex]{graphicx}   
\usepackage{epsfig}             
\usepackage{amsmath}            
\usepackage{amssymb}            
\usepackage{mathtools}
\usepackage{multicol}
\usepackage[bookmarks=true]{hyperref}
\usepackage{xcolor}
\usepackage{siunitx}
\usepackage{lipsum}
\usepackage{scrextend}
\usepackage{algorithm}
\usepackage{algpseudocode}
\usepackage{cite}
\usepackage[inkscapelatex=false]{svg}

\usepackage{bm}

\usepackage{MnSymbol}
\usepackage{mathdots}
\usepackage{amsthm}
\usepackage{empheq}

\input{styles/math}
\input{styles/robotics}

\input{styles/references}

\usepackage{acro}
\usepackage{acro/acro}

\usepackage{siunitx}

\usepackage[caption=false,font=footnotesize]{subfig}

\usepackage{ctable}

\title{
Equivariant IMU Preintegration with Biases:\\a Galilean Group Approach
}

\author{
Giulio Delama$^{1*}$, Alessandro Fornasier$^{1*}$, Robert Mahony$^{2}$ and Stephan Weiss$^{1}$%
\thanks{Manuscript received: July 19, 2024; Accepted November 5, 2024.}%
\thanks{This paper was recommended for publication by Editor Sven Behnke, upon evaluation of the Associate Editor and Reviewers' comments.
This work was supported by the Federal Ministry for Climate Action, Environment, Energy, Mobility, Innovation and Technology (BMK) under the grant agreement 894790 (SALTO) and by the EU-H2020 project BugWright2 (GA 871260).} %
\thanks{$^{1}$Giulio Delama, Alessandro Fornasier, and Stephan Weiss are with the Control of Networked Systems Group, University of Klagenfurt, Austria. $^{*}$G. Delama and A. Fornasier contributed equally. {\tt\footnotesize \{name.surname\}@ieee.org}}%
\thanks{$^{2}$Robert Mahony is with the System Theory and Robotics Lab, Australian National University, Australia. {\tt\footnotesize robert.mahony@anu.edu.au}}%
\thanks{\textbf{{Accepted Nov/2024 for RA-L, DOI:10.1109/LRA.2024.3511424~\copyright IEEE.}}}
}

\begin{document}
\bstctlcite{BSTcontrol}
\markboth{IEEE Robotics and Automation Letters. Preprint Version. Accepted November, 2024}
{Delama \MakeLowercase{\textit{et al.}}: Equivariant IMU Preintegration with Biases: a Galilean Group Approach} 
\maketitle
\input{sections/abstract}
\input{sections/introrelated}
\input{sections/math}

\input{sections/group}
\input{sections/preintegration}
\input{sections/results}
\input{sections/conclusion}

\bibliographystyle{IEEEtran}
\bibliography{bibliography/EqF.bib, bibliography/preintegration.bib, Overleaf/bibliography/extra.bib}
\clearpage
\onecolumn
\appendix
\input{sections/appendix}

\end{document}

%% file: styles/math.tex
\theoremstyle{plain}

\theoremstyle{definition}

\theoremstyle{remark}

\newcommand{\R}{\mathbb{R}}
\newcommand{\norm}[1]{\lVert #1 \rVert}

\newcommand{\set}[2]{\left\{ #1 \;\middle|\; #2 \right\}}

\newcommand{\eye}{\mathbf{I}}

\newcommand{\inv}[1]{{#1}^{-1}}

\newcommand{\grpG}{\mathbf{G}}
\newcommand{\gothg}{\mathfrak{g}}

\newcommand{\liebrk}[2]{\left[#1, #2\right]}
\newcommand{\twoel}[2]{\left(#1,\; #2\right)}
\newcommand{\threeel}[3]{\left(#1,\; #2,\; #3\right)}
\newcommand{\fourel}[4]{\left(#1,\; #2,\; #3,\; #4\right)}

\newcommand{\Adsym}[2]{\mathrm{Ad}_{#1}\left[#2\right]}
\newcommand{\adsym}[2]{\mathrm{ad}_{#1}\left[#2\right]}

\newcommand{\AdMsym}[1]{\mathbf{Ad}^\vee_{#1}}
\newcommand{\adMsym}[1]{\mathbf{ad}^\vee_{#1}}

\newcommand{\SO}[1]{\mathbf{SO}(#1)}
\newcommand{\so}[1]{\mathfrak{so}(#1)}

\newcommand{\SE}[1]{\mathbf{SE}(#1)}

\newcommand{\SEtwo}[1]{\mathbf{SE}_2(#1)}
\newcommand{\setwo}[1]{\mathfrak{se}_2(#1)}

\newcommand{\G}[1]{\mathbf{Gal}(#1)}
\newcommand{\g}[1]{\mathfrak{gal}(#1)}
\newcommand{\elgtsym}[4]{\begin{bmatrix} #1 & #2 & #3 \\ \zeronm{1}{3} & 1 & #4 \\ \zeronm{1}{3} & 0 & 1 \end{bmatrix}}
\newcommand{\hatgtsym}[4]{\begin{bmatrix} #1 & #2 & #3 \\ \zeronm{1}{3} & 0 & #4 \\  \zeronm{1}{3} & 0 & 0 \end{bmatrix}}

\newcommand{\TG}[2]{{#1}_{#2}^\ltimes}

\newcommand{\Jl}[1]{\mathbf{J}_{\mathrm{L}}(#1)}
\newcommand{\Jr}[1]{\mathbf{J}_{\mathrm{R}}(#1)}
\newcommand{\Jlinv}[1]{\mathbf{J}_{\mathrm{L}}(#1)^{-1}}

\newcommand{\J}[2]{\mathbf{\Gamma}_{#1}(#2)}
\newcommand{\Jinv}[2]{\mathbf{\Gamma}_{#1}\!(#2)^{-1}}
\newcommand{\Q}[3]{\mathbf{Q}_{#1}(#2, #3)}
\newcommand{\U}[2]{\mathbf{U}_{#1}(#2)}

%% file: styles/robotics.tex
\newcommand{\Pose}[2]{\prescript{#1}{}{\mathbf{T}}_{#2}}
\newcommand{\Rot}[2]{\prescript{#1}{}{\mathbf{R}}_{#2}}

\newcommand{\Vector}[3]{\prescript{#1}{}{\bm{#2}}_{#3}}

\newcommand{\dotPose}[2]{\dot{\prescript{#1}{}{\mathbf{T}}_{#2}}}
\newcommand{\dotRot}[2]{\dot{\prescript{#1}{}{\mathbf{R}}_{#2}}}
\newcommand{\dotVector}[3]{\prescript{#1}{}{\dot{\bm{#2}}}_{#3}}

\newcommand{\hatVector}[3]{\prescript{#1}{}{\hat{\bm{#2}}}_{#3}}

\newcommand{\curlVector}[3]{\prescript{#1}{}{\tilde{\bm{#2}}}_{#3}}

\newcommand{\ringVector}[3]{\prescript{#1}{}{\mathring{\bm{#2}}}_{#3}}

\newcommand{\vel}[2]{\Vector{#1}{v}{#2}}
\newcommand{\pos}[2]{\Vector{#1}{p}{#2}}
\newcommand{\bias}[2]{\Vector{#1}{b}{#2}}

\newcommand{\dotvel}[2]{\dotVector{#1}{v}{#2}}
\newcommand{\dotpos}[2]{\dotVector{#1}{p}{#2}}
\newcommand{\dotbias}[2]{\dotVector{#1}{b}{#2}}

\newcommand{\hatbias}[2]{\hatVector{#1}{b}{#2}}

\newcommand{\ringbias}[2]{\ringVector{#1}{b}{#2}}

\newcommand{\DelRot}[1]{\Delta\!\Rot{}{#1}}
\newcommand{\Delpos}[1]{\Delta\!\pos{}{#1}}
\newcommand{\Delvel}[1]{\Delta\!\vel{}{#1}}

\newcommand{\Delbiashat}[1]{\Delta\!\hatbias{}{#1}}

\newcommand{\wInp}[2]{\Vector{#1}{w}{#2}}
\newcommand{\tauInp}[2]{\Vector{#1}{\tau}{#2}}
\newcommand{\omegaInp}[2]{\Vector{#1}{\omega}{#2}}
\newcommand{\aInp}[2]{\Vector{#1}{a}{#2}}
\newcommand{\nuInp}[2]{\Vector{#1}{\nu}{#2}}

\newcommand{\curlomegaInp}[2]{\curlVector{#1}{\omega}{#2}}
\newcommand{\curlaInp}[2]{\curlVector{#1}{a}{#2}}
\newcommand{\curlwInp}[2]{\curlVector{#1}{w}{#2}}
\newcommand{\curltauInp}[2]{\curlVector{#1}{\tau}{#2}}
\newcommand{\ringwInp}[2]{\ringVector{#1}{w}{#2}}
\newcommand{\ringtauInp}[2]{\ringVector{#1}{\tau}{#2}}

\newcommand{\Ups}[1]{\mathbf{\Upsilon}_{\!#1}}
\newcommand{\hatUps}[1]{\hat{\mathbf{\Upsilon}}_{\!#1}}
\newcommand{\ringUps}[1]{\mathring{\mathbf{\Upsilon}}_{\!#1}}
\newcommand{\Gam}[1]{\mathbf{\Gamma}_{\!#1}}

\newcommand{\delt}{\delta t}
\newcommand{\Delt}[1]{\Delta t_{#1}}

\newcommand{\sigmad}[1]{\Vector{}{\sigma_{\!\scriptscriptstyle{d}}}{#1}}
\newcommand{\sigmac}[1]{\Vector{}{\sigma_{\!\scriptscriptstyle{c}}}{#1}}

\newcommand{\eyen}[1]{\mathbf{I}_{#1}}

\newcommand{\zeronm}[2]{\mathbf{0}_{#1 \times #2}}

\newcommand{\Rn}[1]{\R^{#1}}

\newcommand{\Rnm}[2]{\R^{#1 \times #2}}

\newcommand{\point}{\;\cdot\;}

%% file: styles/references.tex
\newcommand{\secref}[1]{Sec.~\ref{sec:#1}}

\newcommand{\tabref}[1]{Tab.~\ref{tab:#1}}
\newcommand{\algoref}[1]{Alg.~\ref{alg:#1}}

%% file: sections/abstract.tex
\begin{abstract}
This letter proposes a new approach for Inertial Measurement Unit (IMU) preintegration, a fundamental building block that can be leveraged in different optimization-based Inertial Navigation System (INS) localization solutions.
Inspired by recent advances in equivariant theory applied to biased INSs, we derive a discrete-time formulation of the IMU preintegration on ${\G{3} \ltimes \g{3}}$, the left-trivialization of the tangent group of the Galilean group $\G{3}$.
We define a novel preintegration error that geometrically couples the navigation states and the bias leading to lower linearization error.
Our method improves in consistency compared to existing preintegration approaches which treat IMU biases as a separate state-space.
Extensive validation against state-of-the-art methods, both in simulation and with real-world IMU data, implementation in the Lie++ library, and open-source code are provided.
\end{abstract}


\IEEEpeerreviewmaketitle

%% file: sections/introrelated.tex
\section{Introduction and Related Work}
\IEEEPARstart{I}{nertial} Navigation Systems stand out as localization methods for their ability to utilize data from IMUs and fuse it with other sensors to determine the position and orientation of a mobile robot.
However, classical INS algorithms encounter challenges dealing with biases in the IMU measurements, yielding decreased performances in real-world applications.
The recent introduction of the \emph{equivariant} filter (EqF)~\cite{VanGoor2020EquivariantSpaces, vanGoor2022EquivariantEqF, Fornasier2022EquivariantBiases}, which is a novel and general filter design method for systems evolving on homogeneous spaces, has shown significant improvement in state estimation for biased INSs~\cite{Fornasier2022EquivariantBiases, Fornasier2023EquivariantSystems}.
Researchers have successfully improved consistency, robustness, and accuracy by leveraging \emph{bias-inclusive} symmetries and developing EqFs~\cite{vanGoor2021AnOdometry, VanGoor2023EqVIO:Odometry, Bouazza2023EquivariantMotion, Fornasier2022EquivariantBiases, Fornasier2022OvercomingCalibration, Fornasier2023EquivariantSystems, Scheiber2023RevisitingApproach, Fornasier2023MSCEqF:Navigation} that outperform state-of-the-art methods based on the classical Extended Kalman Filter (EKF) and Invariant Extended Kalman Filter (IEKF)~\cite{7523335}.
Furthermore, the authors in~\cite{Ge2022EquivariantSystems} present an EqF design method for \emph{discrete-time} systems on homogeneous spaces, demonstrating improved convergence and asymptotic performance in simulation with a second-order kinematics system with range and bearing measurements.
Despite advances in the EqF domain, a crucial gap persists in applying this novel and promising theory to optimization-based estimation techniques, whose increasing traction is driven by the growing affordability of powerful and compact computing boards.
In~\cite{Chauchat2018InvariantGroups, Chauchat2022InvariantNoise} the authors propose a nonlinear smoothing algorithm for group-affine observation systems, and in the subsequent work~\cite{Chauchat2023InvariantBiases} they show that utilizing the Two Frames Group (TFG) to better account for IMU biases leads to better performance compared to state-of-the-art methods.
However, research has not yet exploited the potential of equivariant theory and bias-inclusive symmetries~\cite{Fornasier2023EquivariantSystems} applied to optimization-based methods for biased INS, thus presenting an open field for further exploration in robotics.

In the attempt to take a first step in that direction, this work focuses on the \emph{IMU preintegration} problem.
Introduced in~\cite{Lupton2009EfficientInitialization, Lupton2012Visual-inertial-aidedConditions}, IMU preintegration has become an essential component of optimization-based localization methods for INSs as it enables the formulation of a factor between two non-consecutive IMU poses by using inertial measurements only.
Within this context,~\cite{Forster2017On-ManifoldOdometry} marked a significant advance, presenting a novel approach to address the computational complexity of visual-inertial odometry (VIO) by integrating inertial measurements between keyframes into single relative motion constraints.
This fundamental work presents a comprehensive preintegration theory that properly accounts for the rotation group's manifold structure ${\SO{3} \times \Rn{3} \times \Rn{3}}$ and enables efficient computation of Jacobians for optimization.
Later works~\cite{Eckenhoff2019Closed-formNavigation, Yang2020AnalyticNavigation, Tang2022ImpactOptimization} propose novel IMU preintegration models based on the same underlying manifold.

In a recent work~\cite{Brossard2021AssociatingEarth}, it was demonstrated that utilizing the $\SEtwo{3}$ Lie group to encode extended poses and using the \emph{left-invariant} (LI) error definition resulted in significantly improved consistency and accuracy for IMU preintegration with respect to previous methodologies 
that exclusively utilized the $\SO{3}$ Lie group to represent rotations.
This approach represents a substantial advance in IMU preintegration theory as it effectively characterizes \emph{uncertainty propagation} within extended poses, enabling a deeper theoretical description of the problem and ensuring consistency over extended durations.

In another recent study~\cite{Tsao2023AnalyticSystems}, the authors introduce a novel \emph{right-invariant} (RI) IMU preintegration on $\SEtwo{3}$.
The use of a RI parametrization of the error improves the consistency and accuracy of the resulting visual-inertial navigation system (VINS) compared to previous methodologies~\cite{Forster2017On-ManifoldOdometry, Eckenhoff2019Closed-formNavigation, Yang2020AnalyticNavigation} and shows competitiveness against state-of-the-art~\cite{Brossard2021AssociatingEarth}.

A very recent paper~\cite{Wang2024MAVIS:Pre-integration} introduces an enhanced discrete-time IMU preintegration formulation where the mean propagation is based on the exponential function of an automorphism of $\SEtwo{3}$~\cite{vanGoor2023ConstructiveNavigation}, which leads to improved tracking performance, particularly in scenarios with rapid rotational motion.

In~\cite{Fourmy2019AbsoluteMarkers}, the authors recognize the Galilean structure~\cite{Maisser1996Marsden17, DeMontigny2006GalileiContractions} of the preintegration problem and define a novel Lie group called the \emph{IMU deltas matrix Lie group}.
Although they provide a novel recursive calculation for IMU deltas, no coupling with the IMU biases is considered.
In this letter, we generalize these findings and formalize the \emph{Galilean group} $\G{3}$, providing closed-form solutions for all its components.
Similar derivations can be found in the latest report~\cite{Kelly2023AllSGal3}, where $\G{3}$ is referred to as the \emph{Special Galilean Group} $\mathbf{SGal}(3)$.

To the best of the authors' knowledge, previous research on IMU preintegration has not exploited the symmetry of the system to formulate an error that geometrically couples the navigation and the bias states instead of treating them separately.
Encouraged by recent results in EqFs~\cite{Fornasier2022EquivariantBiases, Fornasier2022OvercomingCalibration, Fornasier2023EquivariantSystems, Scheiber2023RevisitingApproach, Fornasier2023MSCEqF:Navigation}, this letter presents a novel equivariant approach to the IMU preintegration problem.
We introduce a novel symmetry based on the left-trivialized tangent group ${\G{3} \ltimes \g{3}}$ and we define a linearized error dynamics based on the \emph{equivariant error}, which effectively establishes a geometric coupling between the navigation states and the bias states, thus resulting in \emph{better linearization} for the navigation states' error.
The advantage of our approach ultimately lies in shifting the linearization error into the bias states, as opposed to retaining it in the more dynamic navigation states~\cite{Fornasier2023EquivariantSystems}.

We extensively tested our novel \emph{equivariant} IMU preintegration both in simulation and with real-world data: our method exhibited superior performance in terms of consistency compared to state-of-the-art preintegration methods~\cite{Forster2017On-ManifoldOdometry, Brossard2021AssociatingEarth, Wang2024MAVIS:Pre-integration, Tsao2023AnalyticSystems}.
In particular, we performed significantly better in all the sequences of the well-known EuRoC MAV dataset~\cite{Burri2016TheDatasets}.

The key contributions of this letter are summarized:
\begin{itemize}
    \item Derivation of a novel discrete-time \emph{equivariant} formulation for the IMU preintegration on the \emph{left-trivialized tangent group} ${\G{3} \ltimes \g{3}}$ that includes IMU biases into the symmetry of the system (summarized in \algoref{preintegration}).
    \item Validation of the proposed approach through an extensive comparison against state-of-the-art IMU preintegration methods in simulation and with real-world data.
    \item Implementation of $\G{3}$ and ${\G{3} \ltimes \g{3}}$ in the publicly available Lie++ library\footnote{https://github.com/aau-cns/Lie-plusplus} and open-sourcing the code, including fast Monte-Carlo batch simulation and real-world IMU datasets evaluation.
\end{itemize}

%% file: sections/math.tex
\section{Notation and Mathematical Preliminaries}
This letter utilizes bold lowercase letters to represent vector quantities, bold capital letters to denote matrices, and regular letters to indicate elements within a symmetry group.
The $n$-dimensional identity matrix is denoted ${\eyen{n} \in \Rnm{n}{n}}$ and the ${n \times m}$ zero matrix is denoted ${\zeronm{n}{m} \in \Rnm{n}{m}}$.

\vspace{-3mm}
\subsection{Lie theory and matrix Lie groups}
A \emph{Lie group} $\grpG$ is a smooth manifold with a smooth group structure.
For any $X, Y \in \grpG$, the group multiplication is denoted by $XY$, while $\inv{X}$ denotes the group inverse and $I$ is the identity element.
The \emph{Lie algebra} $\gothg$ can be modeled as a vector space equivalent to the tangent space at the identity of the group, combined with a bilinear non-associative map $\liebrk{\point}{\point}: \gothg \times \gothg \to \gothg$ called \emph{Lie bracket}.
It is isomorphic to a vector space $\Rn{n}$ where $n=\dim(\gothg)$.
Choosing a basis for $\gothg$ defines two linear mappings between $\gothg$ and $\Rn{n}$ that are known as the \emph{wedge} map and its inverse, the \emph{vee} map:
\begin{equation*}
    (\point)^\wedge: \Rn{n} \to \gothg, \qquad (\point)^\vee: \gothg \to \Rn{n}.
\end{equation*}
The \emph{exponential} map and its inverse, the \emph{logarithmic} map, map elements between $\grpG$ and $\gothg$ in a neighbourhood of the identity in $\grpG$:
\begin{equation*}
    \exp(\point): \gothg \to \grpG, \qquad \log(\point): \grpG \to \gothg.
\end{equation*}
For any $X, Y \in \grpG$, the \emph{left} and \emph{right} translations by $X$ are
\begin{align*}
    \mathrm{L}_X(\point) &: \grpG \to \grpG, \qquad \mathrm{L}_X(Y) = XY, \\
    \mathrm{R}_X(\point) &:  \grpG \to \grpG, \qquad \mathrm{R}_X(Y) = YX.
\end{align*}
For any $X \in \grpG$ and $\Vector{}{u}{}^\wedge \in \gothg$, the big `A' \emph{Adjoint} map for $\grpG$ is
\begin{equation*}
    \Adsym{X}{\point}: \gothg \to \gothg, \qquad \Adsym{X}{\Vector{}{u}{}^\wedge} = \mathrm{dL}_X \circ \mathrm{dR}_{\inv{X}}[\Vector{}{u}{}^\wedge],
\end{equation*}
where $\mathrm{dL}_X[\point]$ and $\mathrm{dR}_X[\point]$ denote the differentials of the left and right translations~\cite{Fornasier2022EquivariantBiases}.
In addition, for any $\Vector{}{u}{}^\wedge, \Vector{}{v}{}^\wedge \in \gothg$, the little `a' \emph{adjoint} map for $\gothg$ is defined as the differential at the identity of $\Adsym{X}{\point}$, and it is equivalent to the Lie bracket:
\begin{equation*}
    \adsym{\Vector{}{u}{}^\wedge}{\point}: \gothg \to \gothg, \qquad \adsym{\Vector{}{u}{}^\wedge}{\Vector{}{v}{}^\wedge} = \liebrk{\Vector{}{u}{}^\wedge}{\Vector{}{v}{}^\wedge}.
\end{equation*}
Considering a Lie group $\grpG$, the G-Torsor is denoted by $\mathcal{G}$ and it represents the set of elements from the Lie group but without the explicit group structure, i.e., the intrinsic manifold.

This study employs \emph{matrix Lie groups}, a subset of Lie groups characterized by elements that can be represented as invertible square matrices, and where the group operation is the matrix multiplication.
Within this context, the big `A' \emph{Adjoint matrix}, denoted $\AdMsym{\mathbf{X}} \in \Rnm{n}{n}$, is defined such that
\begin{equation*}
    \AdMsym{\mathbf{X}}: \Rn{n} \to \Rn{n}, \qquad \AdMsym{\mathbf{X}}\Vector{}{u}{} = (\Adsym{X}{\Vector{}{u}{}^\wedge})^\vee.
\end{equation*}
Similarly, the little `a' \emph{adjoint matrix}, denoted $\adMsym{\Vector{}{u}{}} \in \Rnm{n}{n}$, is defined such that
\begin{equation*}
   \adMsym{\Vector{}{u}{}}\point: \Rn{n} \to \Rn{n}, \qquad\adMsym{\Vector{}{u}{}}\Vector{}{v}{} = \liebrk{\Vector{}{u}{}^\wedge}{\Vector{}{v}{}^\wedge}^\vee.
\end{equation*}
Moreover, for a matrix Lie group $\grpG$, the Adjoint map can be expressed as $\Adsym{X}{\Vector{}{u}{}^\wedge} = \mathbf{X}\Vector{}{u}{}^\wedge\inv{\mathbf{X}}$ and the Lie bracket is equivalent to the matrix commutator, leading to
\begin{equation*}
    \adsym{\Vector{}{u}{}^\wedge}{\Vector{}{v}{}^\wedge} = \liebrk{\Vector{}{u}{}^\wedge}{\Vector{}{v}{}^\wedge} =  \Vector{}{u}{}^\wedge\Vector{}{v}{}^\wedge - \Vector{}{v}{}^\wedge\Vector{}{u}{}^\wedge,
\end{equation*}
where $X \in \grpG$ and $\Vector{}{u}{}^\wedge, \Vector{}{v}{}^\wedge \in \gothg$ are represented as matrices.
It should be noted that the two expressions mentioned above do not hold for products of matrix Lie groups.
For a comprehensive introduction to Lie theory applied to state estimation in robotics, readers can refer to~\cite{Sola2018ARobotics}.

\vspace{-3mm}
\subsection{Semi-direct product and left-trivialized tangent group}
\label{sec:TangentGroup}
Given a Lie group $\grpG$, the tangent space at ${X \in \grpG}$ is associated with the set ${\mathrm{T}_X \grpG = \{ \mathrm{dL}_X[\Vector{}{u}{}^\wedge] = X \Vector{}{u}{}^\wedge \, | \, \Vector{}{u}{}^\wedge \in \gothg\}}$ of left-translated Lie-algebra elements. 
The (left-trivialized) correspondence ${\twoel{X}{\Vector{}{u}{}^\wedge} \mapsto \mathrm{dL}_X[\Vector{}{u}{}^\wedge] \in \mathrm{T}_X \grpG}$ is a ${\grpG \times \gothg}$ parametrization of the tangent bundle. 
This set can be given a \emph{semi-direct product} Lie-group structure~\cite{brockett1972tangent} termed the \emph{left-trivialized tangent group} of $\grpG$ and denoted ${\TG{\grpG}{\gothg} \coloneqq \grpG \ltimes \gothg}$.
Let ${A, B \in \grpG}$ and ${a, b \in \gothg}$ and define ${X = \twoel{A}{a}}$ and ${Y = \twoel{B}{b}}$ as elements of the left-trivialized tangent group ${\grpG \ltimes \gothg}$.
The group operation is the semi-direct product
\vspace{-1mm}
\begin{equation}
    \vspace{-2mm}
    XY = \twoel{AB}{a + \Adsym{A}{b}},
    \label{eq:tgprod}
\end{equation}
the inverse element is
\vspace{-1mm}
\begin{equation}
    \vspace{-2mm}
    \inv{X} = \twoel{\inv{A}}{-\Adsym{\inv{A}}{a}},
    \label{eq:tginv}
\end{equation}
and the identity element is $\twoel{I}{0}$.

The left-trivialized tangent group was first applied to design equivariant filters and observers for INSs in~\cite{Fornasier2022EquivariantBiases, Fornasier2022OvercomingCalibration, Fornasier2023EquivariantSystems, Scheiber2023RevisitingApproach, Fornasier2023MSCEqF:Navigation, Ng2023EquivariantGroups} although the geometric structure was known since the seventies~\cite{brockett1972tangent}.
For a comprehensive introduction to equivariant theory and symmetries for INSs, readers can refer to~\cite{Fornasier2024EquivariantNavigation}.

\vspace{-3mm}
\subsection{Useful matrix Lie groups}
The \emph{special orthogonal} group $\SO{3}$, representing 3D rotations in space, and its Lie algebra $\so{3}$ are defined as
\begin{align*}
    \SO{3} &= \set{\mathbf{A} \in \Rnm{3}{3}}{\mathbf{A}\mathbf{A}^\top = \eye_3,\;\det(\mathbf{A}) = 1}, \\
    \so{3} &= \set{\Vector{}{\omega}{}^\wedge \in \Rnm{3}{3}}{\Vector{}{\omega}{}^\wedge = -\Vector{}{\omega}{}^{\wedge\top}}.
\end{align*}
The \emph{extended special Euclidean} group $\SEtwo{3}$, representing extended poses, and its Lie algebra $\setwo{3}$ are defined as
\begin{align*}
    \SEtwo{3} &= \set{
        \begin{bmatrix}
            \mathbf{A} & \Vector{}{a}{} & \Vector{}{b}{} \\ 
            \zeronm{2}{3} & \multicolumn{2}{c}{\eye_2}
        \end{bmatrix}
        \in \Rnm{5}{5}}
        {\mathbf{A}\in\SO{3},
        \;\Vector{}{a}{},\Vector{}{b}{}\in\Rn{3}} , \\
    \setwo{3} &= \set{
        \begin{bmatrix}
            \Vector{}{\omega}{}^\wedge & \Vector{}{v}{} & \Vector{}{r}{} \\ 
            \zeronm{2}{3} & \multicolumn{2}{c}{\zeronm{2}{2}}
        \end{bmatrix}
        \in \Rnm{5}{5}}
        {\Vector{}{\omega}{}^\wedge\in\so{3},
        \;\Vector{}{v}{},\Vector{}{r}{}\in\Rn{3}} .
\end{align*}

%% file: sections/group.tex
\section{The Galilean Group}
\label{sec:group}
This section presents the \emph{Galilean} matrix Lie group $\G{3}$ and derives closed-form expressions for its components.
$\G{3}$ is the group of 3D rotations, translations in space and time, and transformations between frames of reference that only differ in constant relative motion.
This formalization enables a concise and elegant description of the discrete-time IMU preintegration, and it is fundamental for establishing the \emph{bias-inclusive} symmetry presented in \secref{symmetry}, ultimately improving consistency and minimizing linearization errors~\cite{Fornasier2023EquivariantSystems}.
Furthermore, the analytical closed-form expressions allows fast computation by eliminating the need for numerical or approximated solutions in practical implementation.

Let $X \in \G{3}$ denote an element of the Galilean group, represented in its matrix form as
\begin{equation}
    \mathbf{X} = \elgtsym{\mathbf{A}}{\Vector{}{a}{}}{\Vector{}{b}{}}{c}
    \in \Rnm{5}{5},
    \label{eq:g3elem}
\end{equation}
with $\mathbf{A} \in \SO{3}$, $\Vector{}{a}{}$, $\Vector{}{b}{} \in \Rn{3}$ and $c \in \R$.\\
Let $\Vector{}{x}{} \in \R^{10} $ so that $ \Vector{}{x}{}^\wedge \in \g{3}$ denote an element of the Lie algebra of $\G{3}$, which is represented by the matrix form
\begin{equation}
    \Vector{}{x}{}^\wedge =
    \begin{bmatrix} \Vector{}{\omega}{}\\ \Vector{}{v}{}\\ \Vector{}{r}{}\\ \alpha \end{bmatrix}^{\wedge} = \hatgtsym{\Vector{}{\omega}{}^\wedge}{\Vector{}{v}{}}{\Vector{}{r}{}}{\alpha}
    \in \Rnm{5}{5},
    \label{eq:g3liealg}
\end{equation}
with $\Vector{}{\omega}{}^{\wedge} \in \SO{3}$, $\Vector{}{v}{}, \Vector{}{r}{} \in \Rn{3}$ and $\alpha \in \R$.

\noindent
The inverse element in matrix form is written
\begin{equation}
    \inv{\mathbf{X}} = \elgtsym{\mathbf{A}^\top}{-\mathbf{A}^\top\Vector{}{a}{}}{-\mathbf{A}^\top(\Vector{}{b}{} - c\Vector{}{a}{})}{-c} \in \Rnm{5}{5}.
\end{equation}
The adjoint matrices are defined as
\begin{align}
    \AdMsym{\mathbf{X}} \coloneqq& 
    \begin{bmatrix}
        \mathbf{A} & \zeronm{3}{3} & \zeronm{3}{3} & \zeronm{3}{1}\\
        \Vector{}{a}{}^{\wedge}\mathbf{A} & \mathbf{A} & \zeronm{3}{3} & \zeronm{3}{1}\\
        (\Vector{}{b}{} - c\Vector{}{a}{})^{\wedge}\mathbf{A} & -c\mathbf{A} & \mathbf{A} & \Vector{}{a}{}\\
        \zeronm{1}{3} & \zeronm{1}{3} & \zeronm{1}{3} & 1
    \end{bmatrix},\\
    \adMsym{\Vector{}{x}{}} \coloneqq& 
    \begin{bmatrix}
        \Vector{}{\omega}{}^{\wedge} & \zeronm{3}{3} & \zeronm{3}{3} & \zeronm{3}{1}\\
        \Vector{}{v}{}^{\wedge} & \Vector{}{\omega}{}^{\wedge} & \zeronm{3}{3} & \zeronm{3}{1}\\
        \Vector{}{r}{}^{\wedge} & -\alpha\eye_3 & \Vector{}{\omega}{}^{\wedge} & \Vector{}{v}{}\\
        \zeronm{1}{3} & \zeronm{1}{3} & \zeronm{1}{3} & 0
    \end{bmatrix}.
\end{align}
Closed forms of the exponential and logarithmic maps are
\begin{align}
    \!\!\!\!\exp(\Vector{}{x}{}^{\wedge})\!&=\!
        \elgtsym{\exp(\Vector{}{\omega}{}^{\wedge})}{\J{1}{\Vector{}{\omega}{}}\!\Vector{}{v}{}}{\!\!\J{1}{\Vector{}{\omega}{}}\!\Vector{}{r}{}+\alpha\J{2}{\Vector{}{\omega}{}}\!\Vector{}{v}{}}{\alpha} , 
        \label{eq:expG3}\\
    \!\!\!\!\log(X)\!&=\!
        \hatgtsym{\log(\mathbf{A})}{\!\!\Jinv{1}{\log(\mathbf{A})^{\!\vee}\!}\!\!\Vector{}{a}{}}{\!\!\Jinv{1}{\log(\mathbf{A})^{\!\vee}\!}\mathbf{\Xi}\!\!}{c},
        \label{eq:logG3}
\end{align}
with ${\mathbf{\Xi} = \left(\Vector{}{b}{} - c\,\J{2}{\log(\mathbf{A})^{\vee}}\,\Jinv{1}{\log(\mathbf{A})^{\vee}}\Vector{}{a}{}\right)}$.\\
$\mathbf{\Gamma}_1$ and $\mathbf{\Gamma}_2$ denote respectively the $\SO{3}$ left Jacobian and auxiliary function~\cite{Bloesch2012StateIMU}, both with known closed-form expressions ${\J{1}{\Vector{}{\omega}{}} = \eye_3 + \kappa_1\Vector{}{\omega}{}^{\wedge} + \kappa_2\Vector{}{\omega}{}^{\wedge}\Vector{}{\omega}{}^{\wedge}}$ and ${\J{2}{\Vector{}{\omega}{}} = \frac{1}{2}\eye_3 + \kappa_2\Vector{}{\omega}{}^{\wedge} + \kappa_3\Vector{}{\omega}{}^{\wedge}\Vector{}{\omega}{}^{\wedge}}$ with ${\kappa_1 = \frac{1 - \cos(\norm{\Vector{}{\omega}{}})}{\norm{\Vector{}{\omega}{}}^{2}}}$, ${\kappa_2 = \frac{\norm{\Vector{}{\omega}{}} - \sin(\norm{\Vector{}{\omega}{}})}{\norm{\Vector{}{\omega}{}}^{3}}}$, and ${\kappa_3 = \frac{\norm{\Vector{}{\omega}{}}^{2} + 2\cos(\norm{\Vector{}{\omega}{}}) - 2}{2\norm{\Vector{}{\omega}{}}^{4}}}$.\\
Finally, the closed-form of the $\G{3}$ left Jacobian $\mathbf{J}_{\mathrm{L}}$ is
\begin{equation}
\!\!\!\!\Jl{\Vector{}{x}{}} \!=\! 
\begin{bmatrix}
    \J{1}{\Vector{}{\omega}{}} & \zeronm{3}{3} & \zeronm{3}{3} & \zeronm{3}{1}\\
    \Q{1}{\Vector{}{\omega}{}}{\Vector{}{v}{}} & \J{1}{\Vector{}{\omega}{}} & \zeronm{3}{3} & \zeronm{3}{1}\\
    \mathbf{\Omega}(\Vector{}{\omega}{}, \Vector{}{v}{}, \Vector{}{r}{}, \alpha) & \!\!\!\!-\alpha\U{1}{\Vector{}{\omega}{}} & \!\!\!\J{1}{\Vector{}{\omega}{}} & \!\!\!\J{2}{\Vector{}{\omega}{}}\Vector{}{v}{}\\
    \zeronm{1}{3} & \zeronm{1}{3} & \zeronm{1}{3} & 1\\
\end{bmatrix}\!,
\label{eq:JlG3}
\end{equation}
with $\mathbf{\Omega}(\Vector{}{\omega}{}, \Vector{}{v}{}, \Vector{}{r}{}, \alpha) = \Q{1}{\Vector{}{\omega}{}}{\Vector{}{r}{}}-\alpha\Q{2}{\Vector{}{\omega}{}}{\Vector{}{v}{}}$ and
\begin{align}
    \Q{1}{\Vector{}{\omega}{}}{\Vector{}{z}{}} &= \sum_{p=0}^{\infty}\sum_{k=0}^{\infty}\frac{1}{(p+k+2)!}(\Vector{}{\omega}{}^{\wedge})^{k}\Vector{}{z}{}^{\wedge}(\Vector{}{\omega}{}^{\wedge})^{p} , \\
    \Q{2}{\Vector{}{\omega}{}}{\Vector{}{z}{}} &= \sum_{p=0}^{\infty}\sum_{k=0}^{\infty}\frac{p+1}{(p+k+3)!}(\Vector{}{\omega}{}^{\wedge})^{k}\Vector{}{z}{}^{\wedge}(\Vector{}{\omega}{}^{\wedge})^{p} , \\
    \U{1}{\Vector{}{\omega}{}} &= \sum_{k=0}^{\infty}\frac{1}{(k+2)!}(\Vector{}{\omega}{}^{\wedge})^{k} = \J{1}{\Vector{}{\omega}{}} - \J{2}{\Vector{}{\omega}{}} .
\end{align}
The expressions for $\mathbf{Q}_1$, $\mathbf{Q}_2$, and $\mathbf{U}_1$ are then derived in closed-form, similarly to~\cite{Kelly2023AllSGal3}.
The details of the derivation and the final results are shown in the Appendix~\ref{sec:appendixA}.
An alternative closed-form representation of the left Jacobian in terms of $\mathbf{\Gamma}_1$ and $\mathbf{\Gamma}_2$ is shown in Appendix~\ref{sec:appendixC}.

%% file: sections/preintegration.tex
\section{Equivariant IMU preintegration}
\label{sec:preintegration}
The core contribution of this work is the derivation of a novel \emph{discrete-time} formulation for the \emph{equivariant} IMU preintegration on the left-trivialized tangent group of $\G{3}$, i.e., ${\G{3} \ltimes \g{3}}$.
Specifically, we propose a novel equivariant symmetry for the preintegration problem, which explicitly accounts for the IMU biases, hence improving the propagation of uncertainties and minimizing the linearization errors.

\vspace{-3mm}
\subsection{Biased Inertial Navigation System (INS)}
Consider a mobile robot equipped with an IMU that delivers biased angular velocity and acceleration measurements.
Under non-rotating, flat earth assumption, the noise-free continuous-time biased INS is characterized by the following equations:
\begin{subequations}
\centering
\vspace{-2pt} 
\begin{minipage}{0.25\textwidth}
    \begin{align}
      \dotRot{}{} &= \Rot{}{}\left(\omegaInp{}{} - \bias{}{\omega}\right)^\wedge,\\
      \dotvel{}{} &= \Rot{}{}\left(\aInp{}{} - \bias{}{a}\right) + \Vector{}{g}{},\\
      \dotpos{}{} &= \vel{}{},
    \end{align}
\end{minipage}
\hfill
\begin{minipage}{0.2\textwidth}
    \begin{align}
      \dotbias{}{\omega} &= \tauInp{}{\omega},\\
      \dotbias{}{a} &= \tauInp{}{a},
    \end{align}
\end{minipage}
\vspace{6pt} 
\label{eq:ins}
\end{subequations}
where ${\Rot{}{} \in \SO{3}}$ and ${\vel{}{}, \pos{}{} \in \Rn{3}}$ denote the \emph{core states} (or \emph{navigation states}), i.e., the rigid body orientation, velocity, and position in the global reference frame. ${\bias{}{\omega}, \bias{}{a} \in \Rn{3}}$ denote the \emph{bias state}, ${\Vector{}{g}{} \in \Rn{3}}$ is the gravity vector, and ${\omegaInp{}{}, \aInp{}{} \in \Rn{3}}$ are the \emph{biased} rigid body angular velocity and acceleration. ${\tauInp{}{\omega}, \tauInp{}{a} \in \Rn{3}}$ are inputs used to model the evolution of the bias terms, e.g., they are zero if the biases are modeled as constant quantities.
Similarly to~\cite{Fornasier2022EquivariantBiases,Fornasier2023EquivariantSystems}, by extending \eqref{eq:ins} with additional virtual inputs and bias states, the noise-free biased INS can be reformulated as follows.

Let ${\xi = \twoel{\Pose{}{}}{\bias{}{}} \in \mathcal{M} \coloneqq \mathcal{SE}_2(3) \times \Rn{9}}$ represent the state of the augmented system, where the extended pose ${\Pose{}{} = \threeel{\Rot{}{}}{\vel{}{}}{\pos{}{}} \in \mathcal{SE}_2(3)}$ represents the core states and ${\bias{}{} = \threeel{\bias{}{\omega}}{\bias{}{a}}{\bias{}{\nu}} \in \Rn{9}}$ represents the bias states, including the IMU biases and an additional virtual velocity bias ${\bias{}{\nu} \in \Rn{3}}$, which was initially introduced in~\cite{Fornasier2022EquivariantBiases}.
Define the systems' input ${u = \twoel{\wInp{}{}}{\tauInp{}{}} \in \mathbb{L} \subseteq \Rn{18}}$, where ${\wInp{}{} = \threeel{\omegaInp{}{}}{\aInp{}{}}{\nuInp{}{}} \in \Rn{9}}$ includes the inertial measurements and an additional virtual velocity input ${\nuInp{}{} = \zeronm{3}{1}}$, and ${\tauInp{}{} = \threeel{\tauInp{}{\omega}}{\tauInp{}{a}}{\tauInp{}{\nu}} \in \Rn{9}}$ denotes the biases input.
By defining the matrices
\begin{equation*}
    \mathbf{G}\coloneqq
        \begin{bmatrix}
            \zeronm{3}{3} & \!\!\Vector{}{g}{} & \!\!\zeronm{3}{1}\\ 
            \zeronm{2}{3} & \!\!\zeronm{2}{1} & \!\!\zeronm{2}{1}
        \end{bmatrix}\in\Rnm{5}{5} ,\quad
    \mathbf{N}\coloneqq
        \begin{bmatrix}
            \zeronm{3}{4} & \!\!\zeronm{3}{1} \\ 
            \zeronm{1}{4} & \!\!1\\
            \zeronm{1}{4} & \!\!0
        \end{bmatrix}\in\Rnm{5}{5} ,
\end{equation*}
we can represent the noise-free continuous-time biased INS in compact form as ${\dot\xi = f(\xi, u)}$, that is
\begin{equation}
    \left\{\begin{aligned}
    \dotPose{}{} &= (\mathbf{G} - \mathbf{N})\Pose{}{} + \Pose{}{} (\wInp{}{}^{\wedge} - \bias{}{}^{\wedge} + \mathbf{N}) \\
    \dotbias{}{} &= \tauInp{}{}
    \end{aligned}\right. ,
    \label{eq:system_ct}
\end{equation}
where ${\wInp{}{}^\wedge, \bias{}{}^\wedge \in \setwo{3}}$.

\vspace{-3mm}
\subsection{Discrete-time IMU preintegration}
\label{sec:discrete}
Under the assumption of a constant noise-free input $u_i$ between consecutive time steps $t_i$ and $t_{i+1}$, the exact discretization of \eqref{eq:system_ct} results in the following discrete-time formulation of the noise-free biased INS:
\begin{equation}
    \left\{\begin{aligned}
    \Pose{}{i+1}&=\exp((\mathbf{G} - \mathbf{N})\delt)\Pose{}{i}\exp((\wInp{}{i}^{\wedge} - \bias{}{i}^{\wedge} + \mathbf{N})\delt) \\
    \bias{}{i+1}&=\bias{}{i} + \tauInp{}{i} \delt
    \end{aligned}\right.,
    \label{eq:system_d}
\end{equation}
where ${\xi_i = \twoel{\Pose{}{i}}{\bias{}{i}}}$ and ${u_i = \twoel{\wInp{}{i}}{\tauInp{}{i}}}$ denote the state and the input at the $i$-th time step, and ${\delt = t_{i+1} - t_i}$.
Here, $\exp(\cdot)$ denotes the matrix exponential.
Given two non-consecutive time steps $t_i$ and $t_j$, and defining the \emph{preintegration time} ${\Delt{ij} = t_j - t_i}$, the newest pose $\Pose{}{j}$ is given by
\begin{equation}
    \Pose{}{j} = \Gam{ij}\Pose{}{i}\Ups{ij} , 
    \label{eq:system_ij}
\end{equation}
where $\Gam{ij}$ and $\Ups{ij}$ are exact integration terms defined as
\begin{align}
    \!\!\!\!\!\Gam{ij}\coloneqq
        &\prod_{k=i}^{j-1}\!\exp((\mathbf{G} - \mathbf{N})\delt) =:
        \begin{bmatrix}
            \eye_3&\!\!\Vector{}{g}{}\Delt{ij}&\!\!-\frac{1}{2}\Vector{}{g}{}\Delt{ij}^2\\ 
            \zeronm{1}{3}&\!\!1 &\!\!-\Delt{ij}\\
            \zeronm{1}{3}&\!\!0 &\!\!1
        \end{bmatrix}, \\
    \!\!\!\!\!\Ups{ij}\coloneqq
        &\prod_{k=i}^{j-1}\!\exp((\wInp{}{k}^{\wedge}\!-\!\bias{}{k}^{\wedge}\!+\!\mathbf{N})\delt) =:
        \begin{bmatrix}
            \DelRot{ij}&\!\!\!\Delvel{ij}&\!\!\!\Delpos{ij} \\ 
            \zeronm{1}{3}&\!\!\!1 &\!\!\!\Delt{ij}\\
            \zeronm{1}{3}&\!\!\!0 &\!\!\!1
        \end{bmatrix}.
        \label{eq:upsilon}
\end{align}
Note that $\Gam{ij}$ is not to be confused with $\mathbf{\Gamma}_1$ and $\mathbf{\Gamma}_2$ of \secref{group}.
By reorganizing \eqref{eq:system_ij} we obtain the following expression:
\begin{equation}
    \Ups{ij} = \Pose{}{i}^{-1}\Gam{ij}^{-1}\Pose{}{j} , 
    \label{eq:system_ij_inv}
\end{equation}
with
\begin{equation*}
    \Gam{ij}^{-1} = 
        \begin{bmatrix}
            \eye_3 & -\Vector{}{g}{}\Delt{ij} & -\frac{1}{2}\Vector{}{g}{}\Delt{ij}^2\\ 
            \zeronm{1}{3} & 1 & \Delt{ij}\\
            \zeronm{1}{3} & 0 & 1
        \end{bmatrix} .
\end{equation*}
We refer to $\Ups{ij}$ as the \emph{preintegration matrix}, which stores all the necessary information for relating $\Pose{}{j}$ with $\Pose{}{i}$, as outlined in \eqref{eq:system_ij_inv}.
By reworking \eqref{eq:upsilon}, the iterative computation of the preintegration matrix, starting from ${j=i}$ with ${\Ups{ii} = \eye_5}$, is
\begin{equation}
    \Ups{i(j+1)} = \Ups{ij}\exp((\wInp{}{j}^{\wedge} - \bias{}{j}^{\wedge} + \mathbf{N})\delt) .
    \label{eq:preintegration}
\end{equation}

\vspace{-3mm}
\subsection{Symmetry of the preintegration problem}
\label{sec:symmetry}
The previous equation defines the exact, discrete-time evolution of the preintegration matrix $\Ups{ij}$ as a function of the biased inertial measurements, biases, and time step.
Furthermore, \eqref{eq:upsilon} reveals that the preintegration matrix can be represented as an element of the Galilean group $\G{3}$, which has been formalized in \secref{group}.
This allows us to reformulate the preintegration problem under the lens of equivariance and hence to exploit the structure given by the \emph{equivariant error}~\cite{Mahony2022ObserverEquivariance, vanGoor2022EquivariantEqF, VanGoor2020EquivariantSpaces} to define a linearized error dynamics with reduced linearization error, yielding improved consistency.
From here onward, we adopt a lean notation by defining ${\Ups{k} \coloneqq \Ups{i(i+k)}}$ and ${\bias{}{k} \coloneqq \bias{}{i+k}}$, with ${\xi_0 = \twoel{\Ups{0}}{\bias{}{0}} = \twoel{\eye_5}{\bias{}{i}}}$.

Define ${\mathcal{M} \coloneqq \mathcal{G}al(3) \times \Rn{10}}$ and let ${\xi_k = \twoel{\Ups{k}}{\bias{}{k}} \in \mathcal{M}}$ represent the state of our system.
The preintegration matrix ${\Ups{k} = \fourel{\DelRot{k}}{\Delvel{k}}{\Delpos{k}}{\Delt{k}} \in \mathcal{G}al(3)}$ denotes the pre-integrated navigation states and the preintegration time.
The bias states are denoted by ${\bias{}{k} = \fourel{\bias{}{\omega k}}{\bias{}{ak}}{\bias{}{\nu k}}{b_{\rho k}} \in \Rn{10}}$ and include the IMU biases as well as two additional virtual biases: a velocity bias ${\bias{}{\nu k} \in \Rn{3}}$ and a time bias ${b_{\rho k} \in \R}$.
Define the systems' input ${u_k = \twoel{\wInp{}{k}}{\tauInp{}{k}} \in \mathbb{L} \subseteq \Rn{20}}$, where ${\wInp{}{k} = \fourel{\omegaInp{}{k}}{\aInp{}{k}}{\nuInp{}{k}}{\rho_k} \in \Rn{10}}$ represents the IMU readings and two additional virtual inputs: a velocity input ${\nuInp{}{k} = \zeronm{3}{1}}$, and a time input ${\rho_k = 1}$.
The corresponding bias inputs are ${\tauInp{}{k} = \fourel{\tauInp{}{\omega k}}{\tauInp{}{ak}}{\tauInp{}{\nu k}}{\tau_{\rho k}} \in \Rn{10}}$.\\
The resulting formulation of the noise-free IMU preintegration on the manifold $\mathcal{M}$ is presented as follows:
\begin{equation}
    \left\{\begin{aligned}
    \Ups{k+1} &= \Ups{k}\exp((\wInp{}{k}^{\wedge}-\bias{}{k}^{\wedge})\delt) \\
    \Vector{}{b}{k+1} &= \bias{}{k} + \tauInp{}{k} \delt
    \end{aligned}\right. ,
    \label{eq:manifold_equations}
\end{equation}
where ${\wInp{}{k}^\wedge, \bias{}{k}^\wedge \in \g{3}}$ and $\exp(\cdot)$ denotes the $\G{3}$ group exponential \eqref{eq:expG3}.
The system written in compact form is
\begin{equation}
    \xi_{k+1} = F_{\delt}(\xi_k, u_k) , \quad \xi_0 = (\eye_5, \Vector{}{b}{0}) ,
    \label{eq:manifold_system}
\end{equation}
where $F_{\delt} : \mathcal{M} \times \mathbb{L} \rightarrow \mathcal{M}$ and
\begin{equation*}
    F_{\delt}(\xi_k, u_k) = \twoel{\Ups{k}\exp((\wInp{}{k}^{\wedge}-\bias{}{k}^{\wedge})\delt)}{\bias{}{k} + \tauInp{}{k} \delt} .
\end{equation*}
Having successfully derived the system's evolution on the manifold, our next step is to establish the equivalent (\emph{lifted}) system that evolves within the symmetry group.
This step is crucial, as it enables us to define the \emph{equivariant error} and to subsequently linearize its dynamics, allowing the derivation of the matrices used for the uncertainty propagation while accounting for the input measurement noise.

For the remainder of the letter, we simplify the notation by defining the symmetry group $\grpG$ for the IMU preintegration problem as ${\grpG \coloneqq \G{3} \ltimes \g{3}}$, that is the semi-direct product between $\G{3}$ and its Lie algebra (\secref{TangentGroup}).
Let ${X_k = \twoel{C_k}{\gamma_k} \in \grpG}$ be an element of the symmetry group.
Define the \emph{state action} ${\phi : \grpG \times \mathcal{M} \rightarrow \mathcal{M}}$ as
\begin{equation}
    \phi(X_k, \xi_k) \coloneqq \twoel{\Ups{k}\mathbf{C}_k}{\AdMsym{\mathbf{C}_k^{-1}}(\bias{}{k} - \Vector{}{\gamma}{k}^{\vee})} .
\end{equation}
Then, $\phi$ is a \emph{transitive and free} right group action of $\grpG$ on $\mathcal{M}$.
Fix $\mathring{\xi} \in \mathcal{M}$, then since the action is free, the partial map $\phi_{\mathring{\xi}} : \grpG \to \mathcal{M} $ is a diffeomorphism. 
The inverse \emph{state action} ${\phi^{-1}_{\mathring{\xi}} : \mathcal{M} \rightarrow \grpG}$ is
\begin{equation}
    \phi^{-1}_{\mathring{\xi}}(\xi_k) = \twoel{\mathring{\Ups{}}^{-1}\Ups{k}}{\mathring{\bias{}{}}^{\wedge} - (\AdMsym{\mathring{\Ups{}}^{-1}\Ups{k}}{\bias{}{k}})^{\wedge}} .
\end{equation}
Define the \emph{input action} ${\psi : \grpG \times \mathbb{L} \rightarrow \mathbb{L}}$ as
\begin{equation}
    \psi(X_k, u_k) \coloneqq \twoel{\AdMsym{\mathbf{C}_k^{-1}}(\wInp{}{k} - \Vector{}{\gamma}{k}^{\vee})}{\AdMsym{\mathbf{C}_k^{-1}}\tauInp{}{k}} .
\end{equation}
Then, $\psi$ is a \emph{right group action} of $\grpG$ on $\mathbb{L}$.
The system \eqref{eq:manifold_system} is \emph{equivariant} under the actions ${\phi,\psi}$ of $\grpG$.
The proof is omitted for space limitations but it follows from~\cite{Fornasier2022EquivariantBiases}.

A discrete \emph{lift} for the system is a map ${\Lambda_{\delt} : \mathcal{M} \times \mathbb{L} \rightarrow \grpG}$ with the lift condition ${\phi\left(\Lambda_{\delt}(\xi_k,u_k),\ \xi_k\right) = F_{\delt}(\xi_k, u_k)}$, ${\forall\xi_k\in\mathcal{M}}$ and ${\forall u_k\in\mathbb{L}}$~\cite{Ge2022EquivariantSystems}.
Define the discrete lift $\Lambda_{\delt} : \mathcal{M} \times \mathbb{L} \rightarrow \grpG$ as
\begin{equation}
    \Lambda_{\delt}(\xi_k,u_k) \coloneqq \twoel{\Lambda_{1_{\delt}}(\xi_k,u_k)}{\Lambda_{2_{\delt}}(\xi_k,u_k)} ,
    \vspace{-2mm}
\end{equation}
where
\vspace{-2mm}
\begin{align*}
    \Lambda_{1_{\delt}}(\xi_k,u_k) &= \exp\left(\left(\wInp{}{k}^{\wedge} - \bias{}{k}^{\wedge}\right)\delt\right) , \\
    \Lambda_{2_{\delt}}(\xi_k,u_k) &= \bias{}{k}^{\wedge} - \Adsym{\Lambda_{1_{\delt}}(\xi_k,u_k)}{\bias{}{k}^{\wedge} + \tauInp{}{k}^{\wedge}\delt} .
    \vspace{-1mm}
\end{align*}
Then, $\Lambda_{\delt}$ is an \emph{equivariant lift} for the system in \eqref{eq:manifold_system} with respect to the symmetry group $\grpG$.
Finally, the \emph{lifted system} evolution on $\grpG$ is presented as follows:
\begin{equation}
    X_{k+1} = X_k\Lambda_{\delt}\!\left(\phi_{\mathring{\xi}}(X_k), u_k\right) , \quad X_0 = \phi^{-1}_{\mathring{\xi}}(\xi_0) ,
    \label{eq:lifted_system}
\end{equation}
where ${\mathring{\xi} \in \mathcal{M}}$ is an arbitrarily chosen \emph{state origin} and the group product is defined in \eqref{eq:tgprod}.
Note that we can transfer elements from the symmetry group $\grpG$ to the manifold $\mathcal{M}$ with ${\xi_k = \phi_{\mathring{\xi}}(X_k)}$, and vice versa with ${X_k = \phi^{-1}_{\mathring{\xi}}(\xi_k)}$.

\vspace{-3mm}
\subsection{Linearization of the error dynamics}
The symmetry introduced in the previous subsection allows to exploit the geometric structure of the \emph{equivariant error}~\cite{Mahony2022ObserverEquivariance, vanGoor2022EquivariantEqF, VanGoor2020EquivariantSpaces} and hence to define an error as a measure between the homogeneous space and the symmetry group. Specifically, the equivariant error is defined as follows:
\begin{equation}
    e_k \coloneqq \phi(\hat{X}_k^{-1}, \xi_k) = \twoel{\Ups{k}\hat{\mathbf{C}}_k^{-1}}{\AdMsym{\hat{\mathbf{C}}_k}\bias{}{k} + \hatVector{}{\gamma}{k}^{\vee}} ,
    \label{eq:error}
\end{equation}
where ${\hat{X}_k = \twoel{\hat{C}_k}{\hat{\gamma}_k} \in \grpG}$ denotes the current \emph{state estimate} on the symmetry group, and ${\xi_k = \twoel{\Ups{k}}{\bias{}{k}} \in \mathcal{M}}$ denotes the \emph{actual} (\emph{true}) \emph{state} on the manifold.
Given $e_k \in \mathcal{M}$, we must define a local parametrization in the neighborhood of $\mathring{\xi}$.
A natural choice of parametrization is \emph{logarithmic} or \emph{normal} coordinates.
Let us choose normal coordinates and define a local chart $\vartheta : \mathcal{M} \rightarrow \R^{20}$ as
\begin{equation}
    \vartheta(e_k) \coloneqq \log_G\!\left(\phi_{\mathring{\xi}}^{-1}(e_k)\right) ,
    \label{eq:coords}
\end{equation}
where ${\log_G(\point): \grpG \to \Rn{20}}$ denotes the logarithm of the tangent group ${\G{3} \ltimes \g{3}}$, which is defined as follows:
\begin{equation}
    \log_G\!\left(X_k\right) = \twoel{\log(C_k)^\vee}{\Jlinv{\log(C_k)^\vee}\gamma_k^{\vee}} ,
\end{equation}
with $\log(\cdot)$ and $\Jl{\cdot}$ respectively denoting the $\G{3}$ logarithmic map \eqref{eq:logG3} and the $\G{3}$ left jacobian matrix \eqref{eq:JlG3}.\\
The error \eqref{eq:error} expressed in local coordinates \eqref{eq:coords} is written
\begin{equation}
    \Vector{}{\varepsilon}{k} = \vartheta(e_k) = \log_G\!\left(\phi_{\mathring{\xi}}^{-1}(\phi(\hat{X}_k^{-1}, \xi_k))\right) .
    \label{eq:epsilon}
\end{equation}
By fixing ${\mathring{\xi} = (\ringUps{},\;\ringbias{}{}) = \twoel{\eye_5}{\zeronm{10}{1}}}$ and considering that
\begin{equation}
    \hat{\xi}_k=\phi_{\mathring{\xi}}(\hat{X}_k)=(\ringUps{}\hat{\mathbf{C}}_k,\;\AdMsym{\hat{\mathbf{C}}_k^{-1}}{(\ringbias{}{} - \hatVector{}{\gamma}{k}^\vee)})=(\hatUps{k},\;\hatbias{}{k}) ,
    \label{eq:xi_k}
\end{equation}
we can expand \eqref{eq:epsilon} and express it as
\begin{equation}
    \!\!\!\!\!\!\Vector{}{\varepsilon}{k}\!=\!\twoel{\!\log(\!\Ups{k}\hatUps{k}^{-1}\!)^{\!\vee}\!\!}{\!-\Jlinv{\log(\Ups{k}\hatUps{k}^{-1}\!)^{\!\vee}}\!\AdMsym{\Ups{k}}\!\!({\bias{}{k}}\!-\!\hatbias{}{k}\!)\!}\!.
    \label{eq:error_expanded}
\end{equation}
Let us consider a \emph{noisy} input ${\tilde{u}_k = \twoel{\curlwInp{}{k}}{\curltauInp{}{k}}}$.
${\curlwInp{}{k} = \wInp{}{k} + \Vector{}{\eta}{wk}}$ includes the noisy and biased IMU measurements $\curlomegaInp{}{k}$ and $\curlaInp{}{k}$ as well as the two virtual inputs $\nuInp{}{k}$ and $\rho_k$ introduced in~\secref{symmetry}.
${\curltauInp{}{k} = \tauInp{}{k} + \Vector{}{\eta}{\tau k} = \Vector{}{0}{}\in\Rn{10}}$ denotes the noisy bias input.
The \emph{linearized error dynamics} about ${\Vector{}{\varepsilon}{k} = \Vector{}{0}{}\in\Rn{20}}$ is then derived as follows (for derivation details see Appendix~\ref{sec:appendixB}):
\begin{equation}
    \Vector{}{\varepsilon}{k+1} \approx \hat{\mathbf{A}}_{k+1}\Vector{}{\varepsilon}{k} + \hat{\mathbf{B}}_{k+1}\Vector{}{\eta}{k}
    \label{eq:linearized_dyn}
\end{equation}
with ${\Vector{}{\eta}{k} = \twoel{\Vector{}{\eta}{w k}}{\Vector{}{\eta}{\tau k}} \in \R^{20}}$,
\begin{align}
    \begin{split}
        \hat{\mathbf{A}}_{k+1} 
        &= 
        \begin{bmatrix}
            \eye_{10} && \Jl{\ringwInp{}{k}\delt}\delt \\
            \zeronm{10}{10} && \AdMsym{ \exp(\ringwInp{}{k}^{\wedge}\delt)}
        \end{bmatrix}\label{eq:A} , 
    \end{split}\\
        \hat{\mathbf{B}}_{k+1} &=
        \begin{bmatrix}
            -\Jl{\ringwInp{}{k}\delt}\AdMsym{\hatUps{k}}\delt && \!\!\!\!\zeronm{10}{10} \\
            \zeronm{10}{10} && \!\!\!\!\AdMsym{\hatUps{k+1}} \delt
        \end{bmatrix}\label{eq:B} , 
\end{align}
and $\mathring{u}_k \coloneqq \psi(\hat{X}_k^{-1}, \tilde{u}_k) = \twoel{\ringwInp{}{k}}{\ringtauInp{}{k}} \Rightarrow \ringwInp{}{k} = \AdMsym{\hatUps{k}}(\curlwInp{}{k} - \hatbias{}{k})$.

\vspace{-3mm}
\subsection{Bias update and practical implementation}
From \eqref{eq:manifold_equations} we note that the preintegration matrix $\hatUps{k+1}$ is recursively calculated using the current bias estimate during the mean propagation.
Since it would be very time-consuming to repeat the whole computation every time the estimated bias gets updated during the optimization process, we perform the \emph{bias update} using first-order approximation similarly to~\cite{Tsao2023AnalyticSystems}.
\begin{algorithm}[t]
\caption{IMU Preintegration on $\G{3} \ltimes \g{3}$}
\begin{flushleft}
        \textbf{Define:} ${\mathring{\xi} = \twoel{\eye_5}{\zeronm{10}{1}}}$, ${\hat{X}_0 = \twoel{\eye_5}{-\bias{}{0}^\wedge}}$, ${\mathbf{J_{\xi}}_{0} = \eye_{20}}$, ${\mathbf{\Sigma}_0}$,\\
        ${\qquad\quad\;\mathbf{Q_{\scriptscriptstyle{d}}} = \mathrm{diag}(\sigmad{\omega}^2,\;\sigmad{a}^2,\;\zeronm{4}{1},\;\sigmad{\tau_\omega}^2,\;\sigmad{\tau_a}^2,\;\zeronm{4}{1})}$.\\
        \textbf{Input:} ${\curlwInp{}{k} = \fourel{\curlomegaInp{}{k}}{\curlaInp{}{k}}{\zeronm{3}{1}}{1}}$, ${\tauInp{}{k} = \zeronm{10}{1}}$, \\${\qquad\quad\;\delt = t_{k + 1} - t_k}$.\\
        \textbf{Output:} ${\hat{X}_{k+1} \Leftrightarrow \hat{\xi}_{k+1}}$, ${\mathbf{\Sigma}_{k+1}}$, $\mathbf{J_{\xi}}_{k+1}$.
\end{flushleft}
\begin{algorithmic}[H]
    \Require{IMU measurements $\twoel{\curlomegaInp{}{k}}{\curlaInp{}{k}}$ from $k = 0$ to $N$}
    \For{$k \gets 0 \text{ to } N$}
        \State $\tilde{u}_k \gets \twoel{\curlwInp{}{k}}{\curltauInp{}{k}}$
        \State $\hat{X}_{k+1} \gets \hat{X}_k\Lambda_{\delt}(\phi_{\mathring{\xi}}(\hat{X}_k), \tilde{u}_k)$ \hfill \eqref{eq:lifted_system}
        \State $\hat{\xi}_{k+1} \gets \phi_{\mathring{\xi}}(\hat{X}_{k+1})$ \hfill \eqref{eq:xi_k}
        \State $\mathbf{\Sigma}_{k+1} \gets \hat{\mathbf{A}}_{k+1}\mathbf{\Sigma}_{k}\hat{\mathbf{A}}_{k+1}^\top + \hat{\mathbf{B}}_{k+1}\mathbf{Q_{\scriptscriptstyle{d}}}\hat{\mathbf{B}}_{k+1}^\top$ \hfill \eqref{eq:linearized_dyn}
        \State $\mathbf{J_{\xi}}_{k+1} \gets \mathbf{\Phi_b}_{k+1}\mathbf{J_{\xi}}_{k}$ \hfill \eqref{eq:bias_update}
        \EndFor
\end{algorithmic}
\label{alg:preintegration}
\end{algorithm}
The Jacobian matrix of $\xi_{k+1}$ with respect to $\hatbias{}{k}$ is propagated iteratively as ${\mathbf{J_{\xi}}_{k+1} = \mathbf{\Phi_b}_{k+1}\mathbf{J_{\xi}}_{k}}$ starting from ${\mathbf{J_{\xi}}_{0} = \eye_{20}}$, with
\begin{equation}
    \mathbf{\Phi_b}_{k+1} =
        \begin{bmatrix}
            \eye_{10} && -\AdMsym{\hatUps{k}}\Jl{(\curlwInp{}{k} - \hatbias{}{k})\delt}\delt \\
            \zeronm{10}{10} && \eye_{10}
        \end{bmatrix} .
    \label{eq:bias_update}
\end{equation}
Then, given a new bias estimate ${\hatbias{}{0}^+ \gets \hatbias{}{0} + \Delbiashat{}{}}$, the bias update on the preintegration matrix is performed as
\begin{equation}
    \hatUps{k+1}^+ \approx \exp((\mathbf{J_{\Upsilon}}_{k+1} \Delbiashat{}{})^\wedge)\hatUps{k+1} ,
\end{equation}
where ${\mathbf{J_{\Upsilon}}_{k+1} \in \Rnm{10}{10}}$ is the upper right ${10 \times 10}$ corner of $\mathbf{J_{\xi}}_{k+1}$, and ${\exp(\cdot)}$ is the $\G{3}$ exponential map \eqref{eq:expG3}.

We finally derived all the necessary components for our novel \emph{equivariant} IMU preintegration, summarized in \algoref{preintegration}, which iteratively propagates both mean $\hat{X}_k$ and covariance $\mathbf{\Sigma}_k$ of the equivariant error, together with the Jacobian ${\mathbf{J_{\xi}}_{k}}$ for the bias update.
To streamline C++ development, we added both $\G{3}$ and its left-trivialized tangent group ${\G{3} \ltimes \g{3}}$ to the header-only open-source library Lie++\footnote{https://github.com/aau-cns/Lie-plusplus}.

%% file: sections/results.tex
\section{Experiments and Results}

\begin{table*}[t!]
\centering
\caption{IMU preintegration comparison: NEES median for different preintegration times $\Delt{ij}$ on the EuRoC MAV dataset~\cite{Burri2016TheDatasets}.}
\vspace{-2mm}
\label{tab:euroc}
\begin{tabularx}{\textwidth}{|r| *{3}{>{\centering\arraybackslash}X} | *{3}{>{\centering\arraybackslash}X} | *{3}{>{\centering\arraybackslash}X} | *{3}{>{\centering\arraybackslash}X} | *{3}{>{\centering\arraybackslash}X} |}
\hline
\rule{0pt}{8pt}Method & \multicolumn{3}{c|}{${\SO{3} \times \mathbb{R}^{6} \times \mathbb{R}^{6}}$~\cite{Forster2017On-ManifoldOdometry}} & \multicolumn{3}{c|}{${\text{LI-}\SEtwo{3} \times \mathbb{R}^{6}}$~\cite{Brossard2021AssociatingEarth}} & \multicolumn{3}{c|}{${\text{RI-}\SEtwo{3} \times \mathbb{R}^{6}}$~\cite{Tsao2023AnalyticSystems}} & \multicolumn{3}{c|}{MAVIS~\cite{Wang2024MAVIS:Pre-integration}} & \multicolumn{3}{c|}{${\G{3} \ltimes \g{3}}$} \\
$\scriptscriptstyle{\Delt{ij}}$ & $\scriptscriptstyle{0.2s}$ & $\scriptscriptstyle{0.5s}$ & $\scriptscriptstyle{1.0s}$ & $\scriptscriptstyle{0.2s}$ & $\scriptscriptstyle{0.5s}$ & $\scriptscriptstyle{1.0s}$ & $\scriptscriptstyle{0.2s}$ & $\scriptscriptstyle{0.5s}$ & $\scriptscriptstyle{1.0s}$ & $\scriptscriptstyle{0.2s}$ & $\scriptscriptstyle{0.5s}$ & $\scriptscriptstyle{1.0s}$ & $\scriptscriptstyle{0.2s}$ & $\scriptscriptstyle{0.5s}$ & $\scriptscriptstyle{1.0s}$ \\
\hline
\textbf{MH\_01} & 1.374 & 2.547 & 4.070 & 1.374 & 2.545 & 4.070 & 1.373 & \underline{2.545} & 4.071 & \underline{1.371} & 2.549 & \underline{4.030} & \textbf{1.186} & \textbf{1.866} & \textbf{2.347} \\
\textbf{MH\_02} & 1.113 & 2.077 & 2.871 & 1.113 & 2.077 & 2.871 & 1.114 & 2.078 & 2.870 & \underline{1.096} & \underline{2.071} & \underline{2.867} & \textbf{0.946} & \textbf{1.526} & \textbf{1.755} \\
\textbf{MH\_03} & 1.210 & 2.667 & 4.545 & \underline{1.208} & 2.663 & 4.545 & 1.215 & \underline{2.660} & 4.547& 1.219 & 2.660 & \underline{4.537} & \textbf{1.207} & \textbf{2.269} & \textbf{3.221} \\
\textbf{MH\_04} & 1.222 & 2.638 & \underline{4.506} & \underline{1.216} & 2.641 & 4.534 & 1.220 & \underline{2.635} & 4.531 & 1.220 & 2.636 & 4.524 & \textbf{1.204} & \textbf{2.469} & \textbf{2.757} \\
\textbf{MH\_05} & 1.383 & 3.404 & 4.516 & 1.383 & 3.405 & \underline{4.502} & 1.383 & 3.405 & 4.515 & \underline{1.382} & \underline{3.378} & 4.512 & \textbf{1.377} & \textbf{2.645} & \textbf{2.668} \\
\textbf{V1\_01} & 1.741 & \underline{5.139} & 10.141 & \underline{1.741} & 5.145 & 10.133 & 1.744 & 5.147 & 10.149 & 1.741 & 5.151 & \underline{10.130} & \textbf{1.716} & \textbf{4.837} & \textbf{8.302} \\
\textbf{V1\_02} & \underline{1.363} & \underline{2.456} & 3.844 & 1.370 & 2.466 & \underline{3.834} & 1.371 & 2.463 & 3.837 & 1.382 & 2.467 & 3.841 & \textbf{1.339} & \textbf{2.330} & \textbf{2.458} \\
\textbf{V1\_03} & 1.828 & \underline{3.783} & 5.891 & 1.793 & 3.816 & 5.856 & \underline{1.789} & 3.824 & \underline{5.800} & 1.830 & 3.801 & 6.007 & \textbf{1.734} & \textbf{3.331} & \textbf{3.457} \\
\textbf{V2\_01} & 2.024 & 5.644 & 8.406 & 2.023 & \underline{5.643} & 8.411 & \underline{2.022} & 5.649 & 8.413 & 2.024 & 5.644 & \underline{8.402} & \textbf{2.006} & \textbf{5.350} & \textbf{7.342} \\
\textbf{V2\_02} & 2.579 & 6.627 & 9.788 & \underline{2.574} & \underline{6.620} & 9.573 & 2.579 & 6.627 & 9.761 & 2.576 & 6.634 & \underline{9.732} & \textbf{2.573} & \textbf{6.430} & \textbf{8.422} \\
\textbf{V2\_03} & 3.047 & 8.290 & 13.549 & 3.047 & 8.216 & 13.573 & \underline{3.047} & \underline{8.213} & \underline{13.546} & 3.047 & 8.254 & 13.623 & \textbf{3.025} & \textbf{7.935} & \textbf{11.145} \\
\hline
\multicolumn{16}{c}{The \textbf{best} results are in bold and the \underline{second-best} results are underlined.}
\end{tabularx}
\vspace{-2mm}
\end{table*}

The approach described in the previous section is validated with simulated and real-world data.
A thorough comparison with existing IMU preintegration methodologies is presented.

\vspace{-3mm}
\subsection{Simulation}
In simulation, we generated different analytic trajectories consisting of a circular motion on the $xy$-plane and a cosine wave on the $z$-axis~\cite{Tsao2023AnalyticSystems}.
Synthetic IMU measurements were also generated at a rate of $200~\mathrm{Hz}$.
To validate the proposed approach, we performed several Monte Carlo (MC) simulations varying the trajectory parameters and sampling ${M = 10^3}$ unique realizations of the noise parameters at each time stamp.

We compared the consistency of our method to state-of-the-art IMU preintegration methods~\cite{Forster2017On-ManifoldOdometry, Brossard2021AssociatingEarth, Tsao2023AnalyticSystems, Wang2024MAVIS:Pre-integration} by computing the \emph{Average Normalized Estimation Error Squared} (ANEES)~\cite{Li2012EvaluationTests} as follows:
\vspace{-1mm}
\begin{equation}
    \mathrm{ANEES} \coloneqq \frac{1}{Mn}\sum_{i=1}^{M} \Vector{}{\varepsilon}{i}^\top \mathbf{\Sigma}_i^{-1} \Vector{}{\varepsilon}{i},
    \label{eq:anees}
\end{equation}
where $M$ is the number of MC samples, ${n = \dim(\Vector{}{\varepsilon}{}})$ is the dimension of the error, $\Vector{}{\varepsilon}{i}$ and $\mathbf{\Sigma}_i$ are respectively the error and its covariance for the $i$-th MC realization.
To ensure a rigorous comparison for the \emph{consistency analysis}, we uniformly applied~\eqref{eq:system_ij} for the mean propagation across all methods and we set the error (and covariance) dimension to ${n = 15}$ encompassing the navigation states and the IMU biases.

\begin{figure}[t]
    \centering    \includegraphics[width=1.0\columnwidth]{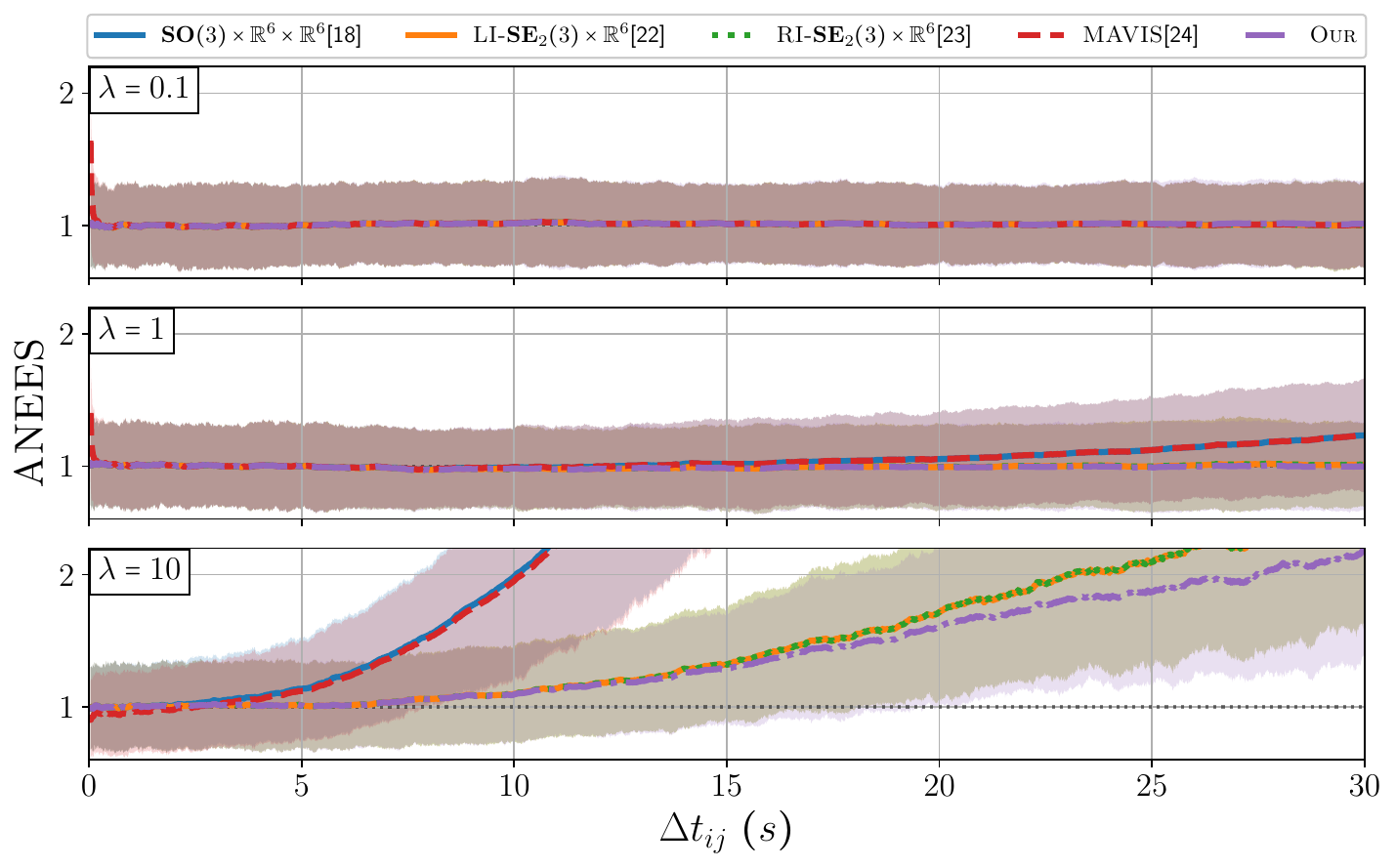}
    \vspace{-6mm}
    \caption{Average NEES for a simulated trajectory with an average speed of ${0.9~\mathrm{m/s}}$ with low noise (top), medium noise (middle), and high noise (bottom).
    The \emph{discrete} noise parameters (${\sigmad{} = \sigmac{}/\sqrt{\Delt{}}}$) are ${\sigmad{\omega} = 7\text{e-}2~\mathrm{rad/s}}$, ${\sigmad{a} = 1.9\text{e-}1~\mathrm{m/s^2}}$, ${\sigmad{\tau_{\omega}} = 1.5\text{e-}4~\mathrm{rad/s^2}}$, ${\sigmad{\tau_{a}} = 1.2\text{e-}2~\mathrm{m/s^3}}$, and $\lambda$ is the noise multiplier.}
    \label{fig:mc_nees}
    \vspace{-2mm}
\end{figure}
\begin{figure}[t]
    \centering    \includegraphics[width=1.0\columnwidth]{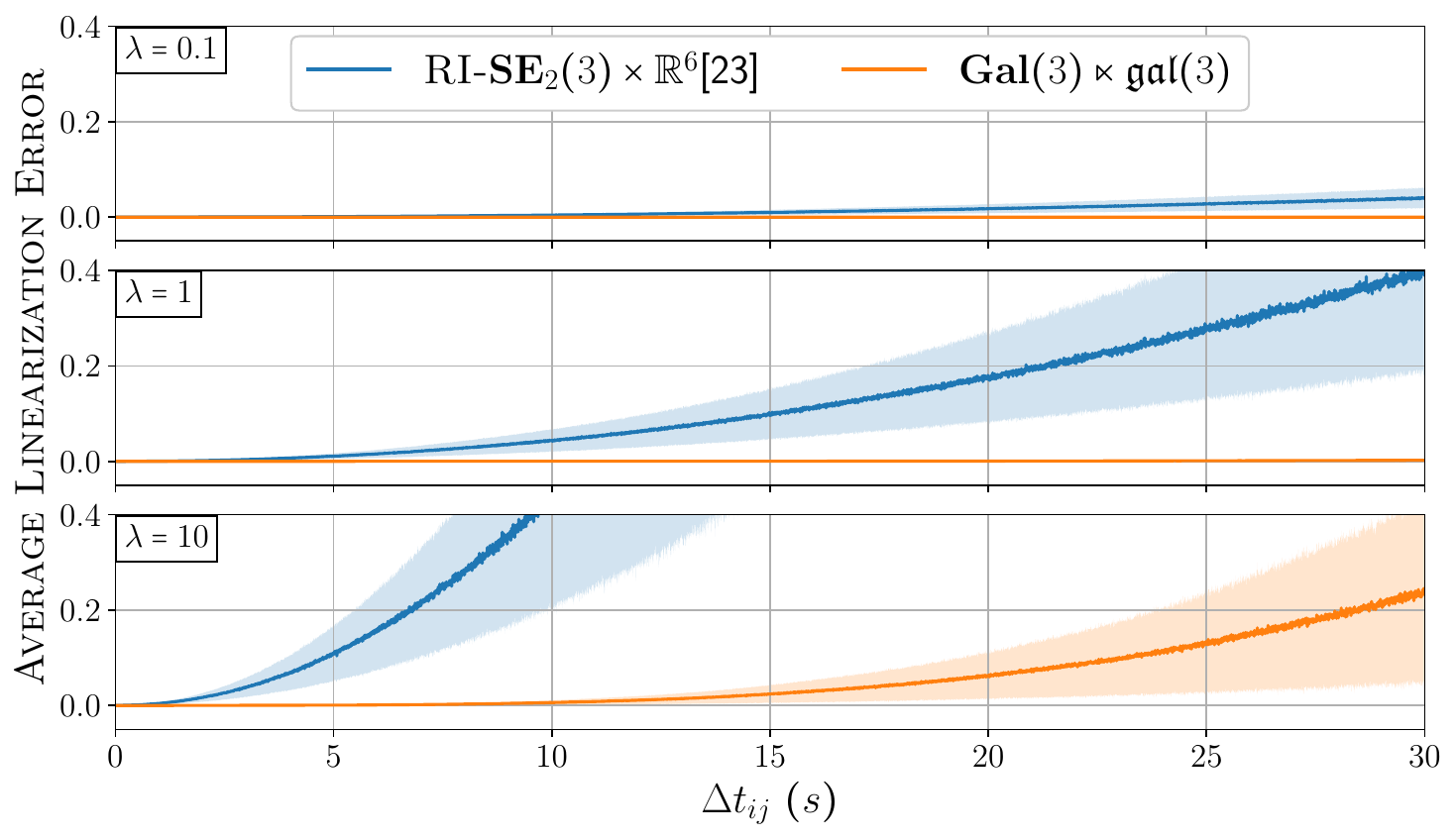}
    \vspace{-6mm}
    \caption{Average linearization error of $\Vector{}{\varepsilon}{\Delta\mathbf{T}_k}$ for a simulated trajectory with an average speed of ${0.9~\mathrm{m/s}}$ with low noise (top), medium noise (middle), and high noise (bottom).
    The \emph{discrete} noise parameters (${\sigmad{} = \sigmac{}/\sqrt{\Delt{}}}$) are ${\sigmad{\omega} = 7\text{e-}2~\mathrm{rad/s}}$, ${\sigmad{a} = 1.9\text{e-}1~\mathrm{m/s^2}}$, ${\sigmad{\tau_{\omega}} = 1.5\text{e-}4~\mathrm{rad/s^2}}$, ${\sigmad{\tau_{a}} = 1.2\text{e-}2~\mathrm{m/s^3}}$, and $\lambda$ is the noise multiplier.}
    \vspace{-2mm}
    \label{fig:mc_err}
\end{figure}

We found that the specific ANEES plot for each MC simulation is influenced by various factors, including the signal-to-noise ratio (SNR), quantization error, and the intensity of the IMU excitation.
Nevertheless, the fundamental pattern remains consistent across the different MC simulations.
Figure~\ref{fig:mc_nees} shows the ANEES for such simulations with varying noise levels.
A parameter $\lambda$ acts as a multiplier for all the discrete noise parameters.
The proposed \emph{equivariant} IMU preintegration on $\G{3} \ltimes \g{3}$ exhibits better consistency over long preintegration times compared to state-of-the-art methods, which do not include the IMU biases into the geometry of the system.

In~\cite{Fornasier2023EquivariantSystems}, the authors demonstrated that exploiting the tangent group of $\SEtwo{3}$ to design an equivariant filter for continuous time biased INSs yields the exact linearization of the navigation error dynamics.
This motivated us to evaluate the \emph{linearization error} of the preintegration problem and to compare our approach against the ${\textsc{RI-}\SEtwo{3} \times \Rn{6}}$~\cite{Tsao2023AnalyticSystems}.
This choice was made due to the common error definition for the preintegrated navigation states $\DelRot{}$, $\Delvel{}$, and $\Delpos{}$ between the two methods while at the same time ${\textsc{RI-}\SEtwo{3} \times \Rn{6}}$ shows best consistency among the tested state-of-the-art approaches.
In fact, the core states' error definition for the ${\textsc{RI-}\SEtwo{3} \times \Rn{6}}$ is ${\Vector{}{\varepsilon}{\Delta\mathbf{T}_k}\!=\!\log(\Delta\mathbf{T}_k \Delta\hat{\mathbf{T}}^{-1}_k)^\vee \in \Rn{9}}$~\cite[(13a)]{Tsao2023AnalyticSystems} where ${\Delta\mathbf{T}_k = \threeel{\DelRot{k}}{\Delvel{k}}{\Delpos{k}} \in \mathcal{SE}_2(3)}$, and $\log(\cdot)$ is the logarithmic map of $\SEtwo{3}$.\\
For the proposed equivariant preintegration, the core states' error is defined according to~\eqref{eq:error_expanded} as ${\Vector{}{\varepsilon}{\Ups{k}}\!\!=\!\log(\Ups{k} \hatUps{k}^{-1})^\vee \in \Rn{10}}$, where ${\Ups{k} = \fourel{\DelRot{k}}{\Delvel{k}}{\Delpos{k}}{\Delt{k}} \in \mathcal{G}al(3)}$ and $\log(\cdot)$ is the logarithmic map of $\G{3}$~\eqref{eq:logG3}.
By assuming $\hat{\Delt{k}}$ to be exact, i.e., ${\hat{\Delt{k}} = \Delt{k}}$, it can be easily proven that ${\Vector{}{\varepsilon}{\Ups{k}} = \twoel{\Vector{}{\varepsilon}{\Delta\mathbf{T}_k}}{0}}$.
This equivalence allows for a direct comparison of the linearization error of $\Vector{}{\varepsilon}{\Delta\mathbf{T}_k}$.
We define the \emph{Average Linearization Error} (ALE) for $\Delta\mathbf{T}_k$ as follows:
\begin{equation}
    \mathrm{ALE} \coloneqq \frac{1}{M}\sum_{i=1}^{M} \norm{\Vector{}{\varepsilon}{\Delta\mathbf{T}_k,i} - \hatVector{}{\varepsilon}{\Delta\mathbf{T}_k,i}},
    \label{eq:ale}
\end{equation}
where $\Vector{}{\varepsilon}{\Delta\mathbf{T}_k}$ denotes the \emph{true error}, and $\hatVector{}{\varepsilon}{\Delta\mathbf{T}_k}$ denotes the \emph{estimated error}.
For the two methods, $\hatVector{}{\varepsilon}{\Delta\mathbf{T}_k}$ is computed via first order propagation from the previous true error and true noise as ${\hatVector{}{\varepsilon}{k} = \hat{\mathbf{A}}_{k+1}\Vector{}{\varepsilon}{k-1} + \hat{\mathbf{B}}_{k+1}\Vector{}{\eta}{k}}$, using~\cite[(6a)]{Tsao2023AnalyticSystems} and~\eqref{eq:linearized_dyn} respectively.
The first 9 elements of $\hatVector{}{\varepsilon}{k}$ represent $\hatVector{}{\varepsilon}{\Delta\mathbf{T}_k}$.

Figure~\ref{fig:mc_err} shows that the equivariant error formulation on ${\G{3} \ltimes \g{3}}$ leads to a significantly lower linearization error compared to the ${\textsc{RI-}\SEtwo{3} \times \Rn{6}}$ method of~\cite{Tsao2023AnalyticSystems}.
The improved linearization error of the proposed methodology can be understood by comparing the structure of the propagation matrix $\hat{\mathbf{A}}_{k+1}$ in~\eqref{eq:A} with the one in~\cite[(6b)]{Tsao2023AnalyticSystems}.
Specifically, the results presented in this work generalize those of~\cite{Fornasier2023EquivariantSystems} to discrete-time inertial navigation systems.
In particular, it can be observed that the $\hat{\mathbf{A}}_{k+1}$ matrix~\eqref{eq:A} relates to the matrix exponential of the continuous-time $\mathbf{A}$ matrix in~\cite[(A.18)]{Fornasier2023EquivariantSystems}.
Therefore, our approach shifts the linearization error from the core navigation states to the bias states yielding a lower linearization error overall.

\begin{figure}[t] 
    \centering
  \subfloat[V1\_03 sequence.\label{fig:euroc_V103_boxplot}\vspace{-3mm}]{%
       \includegraphics[width=1.0\columnwidth]{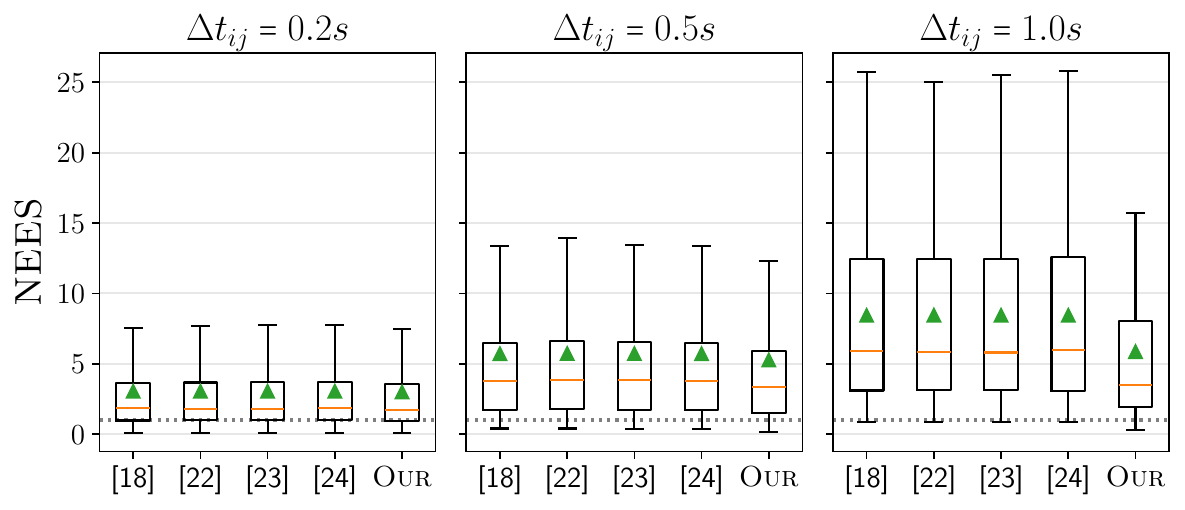}}
    \\
  \subfloat[MH\_01 sequence.\label{fig:euroc_MH01_boxplot}]{%
        \includegraphics[width=1.0\columnwidth]{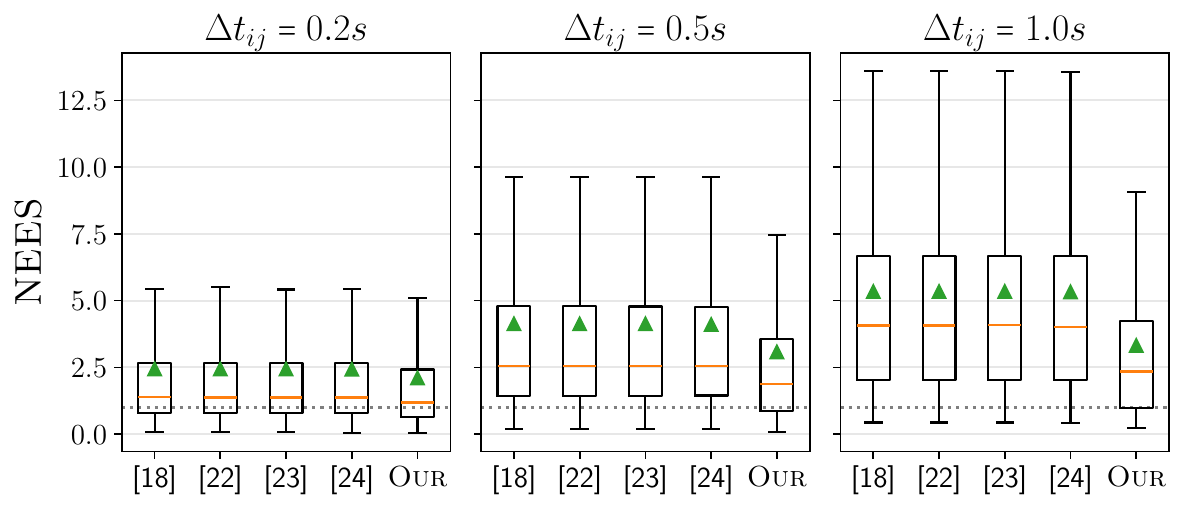}}
  \caption{Consistency comparison on the EuRoC MAV dataset~\cite{Burri2016TheDatasets}: NEES box-plot for different preintegration times ${\Delt{ij}=[0.2s,\; 0.5s,\; 1.0s]}$ on the V1\_03 and MH\_01 sequences.
    Orange lines indicate the medians while green triangles denote the means.
    Longer preintegration times clearly show the benefit of our approach.}
  \label{fig:euroc_boxplot}\vspace{-2mm} 
\end{figure}

\vspace{-3mm}
\subsection{Real-world IMU data}
The simulation results demonstrate noteworthy improvements in both consistency, compared to~\cite{Forster2017On-ManifoldOdometry, Brossard2021AssociatingEarth, Tsao2023AnalyticSystems, Wang2024MAVIS:Pre-integration}, and linearization error, outperforming~\cite{Tsao2023AnalyticSystems}.
Nonetheless, further validation using real-world IMU data becomes imperative due to unmodeled, non-ideal effects, such as non-Gaussian noise distributions, vibration, and aliasing effects, as illustrated in~\cite[Fig. 15]{Brommer2024TheScenarios}.
To assess the real-world performance of our novel \emph{equivariant} IMU preintegration on ${\G{3} \ltimes \g{3}}$, we selected the well-known and widely recognized EuRoC MAV dataset~\cite{Burri2016TheDatasets}, which provides ground-truth poses, velocities and IMU biases, as well as IMU measurements and noise parameters.
We partitioned each dataset sequence into sub-trajectories with a duration corresponding to a preintegration time $\Delt{ij}$.
We then performed the IMU preintegration for each sub-sequence with different methods, starting from the ground-truth values and with the same initial covariance, and we computed the NEES at the end of $\Delt{ij}$.
This process was iterated across various preintegration times, and we analyzed the NEES distribution over all sub-sequences to derive statistical information for each method.
The proposed approach outperforms state-of-the-art methods~\cite{Forster2017On-ManifoldOdometry, Brossard2021AssociatingEarth, Tsao2023AnalyticSystems, Wang2024MAVIS:Pre-integration} in terms of consistency across all sequences in the dataset, as reported in \tabref{euroc}.
Figure~\ref{fig:euroc_boxplot} shows the specific NEES statistics for the V1\_03 and the MH\_01 sequences.

%% file: sections/conclusion.tex
\section{Conclusion}
This letter introduces a theoretical framework for IMU preintegration on ${\G{3} \ltimes \g{3}}$, i.e., the left-trivialized tangent group of the Galilean group $\G{3}$, and successfully demonstrates its performance when compared with state-of-the-art methodologies.
Specifically, we leverage an equivariant symmetry to define a novel preintegration error that geometrically couples the navigation states and the bias states, ultimately improving the covariance propagation, and hence the consistency, for the preintegrated IMU measurements.
An extensive validation against existing IMU preintegration methods is carried out, both in simulation and with real-world datasets.
Results show that the proposed approach not only achieves lower linearization error than state-of-the-art but it also achieves the best NEES in every sequence of the EuRoC MAV dataset and for every preintegration time.
To facilitate integration, reproducibility, and further comparison we open-source our code, including a practical implementation of the symmetry groups into the Lie++ library.

%% file: sections/appendix.tex
\subsection{Closed-form derivations for $\G{3}$ left Jacobian matrix}
\label{sec:appendixA}
This section presents the comprehensive derivation of the exact closed-form expressions for $\mathbf{Q}_1$, $\mathbf{Q}_2$, and $\mathbf{U}_1$ introduced in~\secref{group}.
We currently include only the final expressions, with a full step-by-step derivation to be provided in future versions.

\begin{equation}
    \begin{split}
    \Q{1}{\Vector{}{\omega}{}}{\Vector{}{z}{}} &\coloneqq \sum_{p=0}^{\infty}\sum_{k=0}^{\infty}\frac{1}{(p+k+2)!}(\Vector{}{\omega}{}^{\wedge})^{k}\Vector{}{z}{}^{\wedge}(\Vector{}{\omega}{}^{\wedge})^{p} = \\
    & = \frac{1}{2}\Vector{}{z}{}^{\wedge} + \left(\frac{\norm{\Vector{}{\omega}{}} - \sin(\norm{\Vector{}{\omega}{}})}{\norm{\Vector{}{\omega}{}}^{3}}\right)\left(\Vector{}{\omega}{}^{\wedge}\Vector{}{z}{}^{\wedge} + \Vector{}{z}{}^{\wedge}\Vector{}{\omega}{}^{\wedge} + \Vector{}{\omega}{}^{\wedge}\Vector{}{z}{}^{\wedge}\Vector{}{\omega}{}^{\wedge}\right) + \\
    & + \left(\frac{\norm{\Vector{}{\omega}{}}^{2} + 2\cos(\norm{\Vector{}{\omega}{}}) - 2}{2\norm{\Vector{}{\omega}{}}^{4}}\right)\left(\Vector{}{\omega}{}^{\wedge}\Vector{}{\omega}{}^{\wedge}\Vector{}{z}{}^{\wedge} + \Vector{}{z}{}^{\wedge}\Vector{}{\omega}{}^{\wedge}\Vector{}{\omega}{}^{\wedge} - 3\Vector{}{\omega}{}^{\wedge}\Vector{}{z}{}^{\wedge}\Vector{}{\omega}{}^{\wedge}\right) + \\
    & + \left(\frac{2\norm{\Vector{}{\omega}{}} - 3\sin(\norm{\Vector{}{\omega}{}}) + \norm{\Vector{}{\omega}{}}\cos(\norm{\Vector{}{\omega}{}})}{\norm{\Vector{}{\omega}{}}^{5}}\right)\Vector{}{\omega}{}^{\wedge}\Vector{}{\omega}{}^{\wedge}\Vector{}{z}{}^{\wedge}\Vector{}{\omega}{}^{\wedge} ,
    \end{split}
\end{equation}
\begin{equation} 
    \begin{split}   
    \Q{2}{\Vector{}{\omega}{}}{\Vector{}{z}{}} &\coloneqq \sum_{p=0}^{\infty}\sum_{k=0}^{\infty}\frac{p+1}{(p+k+3)!}(\Vector{}{\omega}{}^{\wedge})^{k}\Vector{}{z}{}^{\wedge}(\Vector{}{\omega}{}^{\wedge})^{p} = \\
    & = \frac{1}{6}\Vector{}{z}{}^{\wedge} + \left(\frac{\norm{\Vector{}{\omega}{}}^{2} + 2\cos(\norm{\Vector{}{\omega}{}}) - 2}{2\norm{\Vector{}{\omega}{}}^{4}}\right)\Vector{}{z}{}^{\wedge}\Vector{}{\omega}{}^{\wedge} + \\
    & + \left(\frac{-2\cos(\norm{\Vector{}{\omega}{}}) - \norm{\Vector{}{\omega}{}}\sin(\norm{\Vector{}{\omega}{}}) + 2}{\norm{\Vector{}{\omega}{}}^{4}}\right)\Vector{}{\omega}{}^{\wedge}\Vector{}{z}{}^{\wedge} + \\
    & + \left(\frac{\norm{\Vector{}{\omega}{}}^{3} - 6\norm{\Vector{}{\omega}{}} + 6\sin(\norm{\Vector{}{\omega}{}})}{6\norm{\Vector{}{\omega}{}}^{5}}\right)\Vector{}{z}{}^{\wedge}\Vector{}{\omega}{}^{\wedge}\Vector{}{\omega}{}^{\wedge} + \\
    & + \left(\frac{\norm{\Vector{}{\omega}{}}^{3} + 6\norm{\Vector{}{\omega}{}}\cos(\norm{\Vector{}{\omega}{}}) + 6\norm{\Vector{}{\omega}{}} - 12\sin(\norm{\Vector{}{\omega}{}})}{6\norm{\Vector{}{\omega}{}}^{5}}\right)\Vector{}{\omega}{}^{\wedge}\Vector{}{\omega}{}^{\wedge}\Vector{}{z}{}^{\wedge} + \\
    & + \left(\frac{-\norm{\Vector{}{\omega}{}}^{3} - 12\norm{\Vector{}{\omega}{}}\cos(\norm{\Vector{}{\omega}{}}) - 3\norm{\Vector{}{\omega}{}}^{2}\sin(\norm{\Vector{}{\omega}{}}) + 12\sin(\norm{\Vector{}{\omega}{}})}{6\norm{\Vector{}{\omega}{}}^{5}}\right)\Vector{}{\omega}{}^{\wedge}\Vector{}{z}{}^{\wedge}\Vector{}{\omega}{}^{\wedge} + \\
    & + \left(\frac{\norm{\Vector{}{\omega}{}}^{2} + \norm{\Vector{}{\omega}{}}^{2}\cos(\norm{\Vector{}{\omega}{}}) - 4\norm{\Vector{}{\omega}{}}\sin(\norm{\Vector{}{\omega}{}}) - 4\cos(\norm{\Vector{}{\omega}{}}) + 4}{2\norm{\Vector{}{\omega}{}}^{6}}\right)\Vector{}{\omega}{}^{\wedge}\Vector{}{\omega}{}^{\wedge}\Vector{}{z}{}^{\wedge}\Vector{}{\omega}{}^{\wedge} ,
    \end{split}
\end{equation}
\begin{equation}  
    \begin{split} 
    \mathbf{U}_1(\Vector{}{\omega}{}) &\coloneqq \sum_{k=0}^{\infty}\frac{1}{(k+2)!}(\Vector{}{\omega}{}^{\wedge})^{k} = \\
    & = \frac{1}{2}\eye_3 + \left(\frac{\sin(\norm{\Vector{}{\omega}{}}) - \norm{\Vector{}{\omega}{}}\cos(\norm{\Vector{}{\omega}{}})}{\norm{\Vector{}{\omega}{}}^{3}}\right)\Vector{}{\omega}{}^{\wedge} + \\
    & + \left(\frac{\norm{\Vector{}{\omega}{}}^{2} - 2\norm{\Vector{}{\omega}{}}\sin(\norm{\Vector{}{\omega}{}}) - 2\cos(\norm{\Vector{}{\omega}{}}) + 2}{2\norm{\Vector{}{\omega}{}}^{4}}\right)\Vector{}{\omega}{}^{\wedge}\Vector{}{\omega}{}^{\wedge} = \\
    & = \left(\frac{\norm{\Vector{}{\omega}{}}\sin(\norm{\Vector{}{\omega}{}}) + \cos(\norm{\Vector{}{\omega}{}}) - 1}{\norm{\Vector{}{\omega}{}}^{2}}\right)\eye_3 + \left(\frac{\sin(\norm{\Vector{}{\omega}{}}) - \norm{\Vector{}{\omega}{}}\cos(\norm{\Vector{}{\omega}{}})}{\norm{\Vector{}{\omega}{}}^{3}}\right)\Vector{}{k}{}^{\wedge} + \\
    & + \left(0.5 - \frac{\norm{\Vector{}{\omega}{}}\sin(\norm{\Vector{}{\omega}{}}) + \cos(\norm{\Vector{}{\omega}{}}) - 1}{\norm{\Vector{}{\omega}{}}^{2}}\right)\Vector{}{k}{}\Vector{}{k}{}^{\top} \\
    & = \J{1}{\Vector{}{\omega}{}} - \J{2}{\Vector{}{\omega}{}},
    \end{split}
\end{equation}
where ${\Vector{}{k}{} = \frac{\Vector{}{\omega}{}}{\norm{\Vector{}{\omega}{}}}}$.
Note that ${\Vector{}{\omega}{}^{\wedge}\Vector{}{\omega}{}^{\wedge}\Vector{}{z}{}^{\wedge}\Vector{}{\omega}{}^{\wedge}\Vector{}{\omega}{}^{\wedge} = -\norm{\Vector{}{\omega}{}}^2\Vector{}{\omega}{}^{\wedge}\Vector{}{z}{}^{\wedge}\Vector{}{\omega}{}^{\wedge}}$ and ${\Vector{}{\omega}{}^{\wedge}\Vector{}{\omega}{}^{\wedge}\Vector{}{z}{}^{\wedge}\Vector{}{\omega}{}^{\wedge} = \Vector{}{\omega}{}^{\wedge}\Vector{}{z}{}^{\wedge}\Vector{}{\omega}{}^{\wedge}\Vector{}{\omega}{}^{\wedge}}$.

\subsection{Derivation of the error linearization matrices}
\label{sec:appendixB}
This section presents a step-by-step derivation of the linearization matrices $\hat{\mathbf{A}}_{k+1}$ and $\hat{\mathbf{B}}_{k+1}$ in~\eqref{eq:linearized_dyn}.
Consider the error expressed in local coordinates $\Vector{}{\varepsilon}{k+1} = \twoel{\Vector{}{\varepsilon}{\Upsilon_{k+1}}}{\Vector{}{\varepsilon}{b_{k+1}}}$ as in~\eqref{eq:error_expanded}.
One can write:
\begin{align*}
    \Vector{}{\varepsilon}{\Upsilon_{k+1}} =& \log\left(\Ups{k+1}\hatUps{k+1}^{-1}\right)^{\vee} \\
    =& \log\left(\Ups{k}\exp((\wInp{}{k}-\bias{}{k})^\wedge\delt)\exp((\curlwInp{}{k}-\hatbias{}{k})^\wedge\delt)^{-1}\hatUps{k}^{-1}\right)^{\vee} \\
    =& \log\left(\exp(\Vector{}{\varepsilon}{\Upsilon_{k}}^\wedge)\hatUps{k}\exp((\curlwInp{}{k} - \Vector{}{\eta}{w k} -\hatbias{}{k} + \AdMsym{\hatUps{k}^{-1}}\Vector{}{\varepsilon}{b_{k}})^\wedge\delta t)\exp((\curlwInp{}{k}-\hatbias{}{k})^{\wedge}\delt)^{-1}\hatUps{k}^{-1}\right)^{\vee} \\
    =& \log\left(\exp(\Vector{}{\varepsilon}{\Upsilon_{k}}^{\wedge})\hatUps{k}\exp((\Jl{(\curlwInp{}{k}-\hatbias{}{k})\delt}(\AdMsym{\hatUps{k}^{-1}}\Vector{}{\varepsilon}{b_{k}} - \Vector{}{\eta}{w k}))^\wedge\delt)\hatUps{k}^{-1}\right)^{\vee} \\
    =& \log\left(\exp(\Vector{}{\varepsilon}{\Upsilon_{k}}^{\wedge})\exp((\AdMsym{\hatUps{k}}\Jl{(\curlwInp{}{k}-\hatbias{}{k})\delt}(\AdMsym{\hatUps{k}^{-1}}\Vector{}{\varepsilon}{b_{k}} - \Vector{}{\eta}{w k}))^\wedge\delt)\right)^{\vee} \\
    \approx& \Vector{}{\varepsilon}{\Upsilon_{k}} + \AdMsym{\hatUps{k}}\Jl{(\curlwInp{}{k}-\hatbias{}{k})\delt}\AdMsym{\hatUps{k}^{-1}}\delt\Vector{}{\varepsilon}{b_{k}} - \AdMsym{\hatUps{k}}\Jl{(\curlwInp{}{k}-\hatbias{}{k})\delt}\delt\Vector{}{\eta}{w k} \\
    \approx& \Vector{}{\varepsilon}{\Upsilon_{k}} + \Jl{\AdMsym{\hatUps{k}}(\curlwInp{}{k}-\hatbias{}{k})\delt}\delt\Vector{}{\varepsilon}{b_{k}} - \AdMsym{\hatUps{k}}\Jl{(\curlwInp{}{k}-\hatbias{}{k})\delt}\delt\Vector{}{\eta}{w k} \\
    \approx& \Vector{}{\varepsilon}{\Upsilon_{k}} + \Jl{\ringwInp{}{k}\delt}\delt\Vector{}{\varepsilon}{b_{k}} - \Jl{\ringwInp{}{k}\delt}\AdMsym{\hatUps{k}}\delt\Vector{}{\eta}{w k} \quad , \\
    \Vector{}{\varepsilon}{b_{k+1}} =& -\Jlinv{\Vector{}{\varepsilon}{\Upsilon_{k+1}}}\AdMsym{\Ups{k+1}}(\bias{}{k+1} - \hatbias{}{k+1}) \\
    \approx& -\AdMsym{\exp(\Vector{}{\varepsilon}{\Upsilon_{k+1}}^\wedge)\hatUps{k+1}}(\bias{}{k} + \tauInp{}{k}\delt - \hatbias{}{k}) \\
    \approx& -\AdMsym{\exp(\Vector{}{\varepsilon}{\Upsilon_{k+1}}^\wedge)\hatUps{k+1}}(\hatbias{}{k} - \AdMsym{\hatUps{k}^{-1}}\Vector{}{\varepsilon}{b_{k}} - \Vector{}{\eta}{\tau k}\delt - \hatbias{}{k}) \\
    \approx& \AdMsym{\hatUps{k+1}}(\AdMsym{\hatUps{k}^{-1}}\Vector{}{\varepsilon}{b_{k}} + \Vector{}{\eta}{\tau k}\delt) \\
    \approx& \AdMsym{\hatUps{k}\exp((\curlwInp{}{k}-\hatbias{}{k})^{\wedge}\delt)\hatUps{k}^{-1}}\Vector{}{\varepsilon}{b_{k}} + \AdMsym{\hatUps{k+1}}\delt\Vector{}{\eta}{\tau k} \\
    \approx& \AdMsym{\exp((\AdMsym{\hatUps{k}}(\curlwInp{}{k}-\hatbias{}{k}))^{\wedge}\delt)}\Vector{}{\varepsilon}{b_{k}} + \AdMsym{\hatUps{k+1}}\delt\Vector{}{\eta}{\tau k} \\
    \approx& \AdMsym{\exp(\ringwInp{}{k}^\wedge\delt)}\Vector{}{\varepsilon}{b_{k}} + \AdMsym{\hatUps{k+1}}\delt\Vector{}{\eta}{\tau k} \quad ,
\end{align*}
which is equivalent to writing $\Vector{}{\varepsilon}{k+1} \approx \hat{\mathbf{A}}_{k+1}\Vector{}{\varepsilon}{k} + \hat{\mathbf{B}}_{k+1}\Vector{}{\eta}{k}$ 
with
\begin{equation*}
        \hat{\mathbf{A}}_{k+1} = 
        \begin{bmatrix}
            \eye_{10} && \Jl{\ringwInp{}{k}\delt}\delt \\
            \zeronm{10}{10} && \AdMsym{ \exp(\ringwInp{}{k}^{\wedge}\delt)}
        \end{bmatrix} , \qquad
        \hat{\mathbf{B}}_{k+1} =
        \begin{bmatrix}
            -\Jl{\ringwInp{}{k}\delt}\AdMsym{\hatUps{k}}\delt && \zeronm{10}{10} \\
            \zeronm{10}{10} && \AdMsym{\hatUps{k+1}} \delt
        \end{bmatrix} .
\end{equation*}

\subsection{Alternative formulation of the left and right Jacobians for $\SE{3}$, $\SEtwo{3}$, and $\G{3}$}
\label{sec:appendixC}
Let $\Vector{}{x}{}\in \Rn{n}$ such that $\Vector{}{x}{}^\wedge \in \gothg$ is an element of a Lie algebra of dimension $n$.
The left and right Jacobian matrices $\Jl{\Vector{}{x}{}}, \Jr{\Vector{}{x}{}} \in \Rnm{n}{n}$ satisfy the following approximations for small $\delta\!\Vector{}{x}{} \in \Rn{n}$:
\begin{equation}
    \exp((\Vector{}{x}{} + \delta\!\Vector{}{x}{})^\wedge) \approx \exp((\Jl{\Vector{}{x}{}}\delta\!\Vector{}{x}{})^\wedge) \exp(\Vector{}{x}{}^\wedge) \approx \exp(\Vector{}{x}{}^\wedge) \exp((\Jr{\Vector{}{x}{}}\delta\!\Vector{}{x}{})^\wedge) .
\end{equation}

\subsubsection{$\SE{3}$}
Results for $\SE{3}$ can be directly drawn from the one of $\SEtwo{3}$. 

\subsubsection{$\SEtwo{3}$}
Let $\Vector{}{x}{} = \threeel{\Vector{}{\omega}{}}{\Vector{}{v}{}}{\Vector{}{r}{}} \in \Rn{9}$ such that $\Vector{}{x}{}^\wedge \in \setwo{3}$.
The exponential map of $\SEtwo{3}$ is defined as follows:
\begin{equation}
    \exp(\Vector{}{x}{}^{\wedge}) = 
        \elgtsym{\exp(\Vector{}{\omega}{}^{\wedge})}{\J{1}{\Vector{}{\omega}{}}\Vector{}{v}{}}{\J{1}{\Vector{}{\omega}{}}\Vector{}{r}{}}{0} .
\end{equation}
The $\SEtwo{3}$ left Jacobian matrix $\Jl{\Vector{}{x}{}} \in \Rnm{9}{9}$ can be expressed in the following closed-form
\begin{equation}
\Jl{\Vector{}{x}{}} = 
\begin{bmatrix}
    \J{1}{\Vector{}{\omega}{}} & \zeronm{3}{3} & \zeronm{3}{3} \\
    \Q{1}{\Vector{}{\omega}{}}{\Vector{}{v}{}} & \J{1}{\Vector{}{\omega}{}} & \zeronm{3}{3} \\
    \Q{1}{\Vector{}{\omega}{}}{\Vector{}{r}{}} & \zeronm{3}{3} & \J{1}{\Vector{}{\omega}{}}
\end{bmatrix} = 
\begin{bmatrix}
    \J{1}{\Vector{}{\omega}{}} & \zeronm{3}{3} & \zeronm{3}{3}\\
    H_1(\Vector{}{\omega}{}, \Vector{}{v}{}) + (\J{1}{\Vector{}{\omega}{}}\Vector{}{v}{})^\wedge\J{1}{\Vector{}{\omega}{}} & \J{1}{\Vector{}{\omega}{}} & \zeronm{3}{3}\\
    H_1(\Vector{}{\omega}{}, \Vector{}{r}{}) + (\J{1}{\Vector{}{\omega}{}}\Vector{}{r}{})^\wedge\J{1}{\Vector{}{\omega}{}} & \zeronm{3}{3} & \J{1}{\Vector{}{\omega}{}}
\end{bmatrix} ,
\end{equation}
where we define the derivatives
\begin{align}
    H_1(\Vector{}{\omega}{}, \Vector{}{z}{}) \coloneqq \frac{\partial}{\partial \Vector{}{\omega}{}}(\J{1}{\Vector{}{\omega}{}}\Vector{}{z}{}) &= - \kappa_4 \Vector{}{\omega}{}^\wedge\Vector{}{z}{}\Vector{}{\omega}{}^\top - \kappa_1 \Vector{}{z}{}^\wedge - \kappa_5 \Vector{}{\omega}{}^\wedge\Vector{}{\omega}{}^\wedge\Vector{}{z}{}\Vector{}{\omega}{}^\top + \kappa_2 H_0(\Vector{}{\omega}{}, \Vector{}{z}{}) , \label{eq:h1}\\
    H_0(\Vector{}{\omega}{}, \Vector{}{z}{}) \coloneqq \frac{\partial}{\partial \Vector{}{\omega}{}}(\Vector{}{\omega}{}^\wedge\Vector{}{\omega}{}^\wedge\Vector{}{z}{}) &= \Vector{}{z}{}^\top\Vector{}{\omega}{} \eye_3 + \Vector{}{\omega}{}\Vector{}{z}{}^\top - 2\Vector{}{z}{}\Vector{}{\omega}{}^\top , \label{eq:h0}
\end{align}
and the scalar coefficients
\begin{align}
    &\kappa_0 = \frac{\sin(\norm{\Vector{}{\omega}{}})}{\norm{\Vector{}{\omega}{}}}, &&\kappa_1 = \frac{1 - \cos(\norm{\Vector{}{\omega}{}})}{\norm{\Vector{}{\omega}{}}^2}, &&&\kappa_2 = \frac{1 - \kappa_0}{\norm{\Vector{}{\omega}{}}^2}, &&&&\kappa_4 = \frac{2\kappa_1 - \kappa_0}{\norm{\Vector{}{\omega}{}}^2}, &&&&&\kappa_5 = \frac{3\kappa_2 - \kappa_1}{\norm{\Vector{}{\omega}{}}^2} . \label{eq:k01245}
\end{align}
Then we can write $\Q{1}{\Vector{}{\omega}{}}{\Vector{}{z}{}}$ as a function of $H_1(\Vector{}{\omega}{}, \Vector{}{z}{})$ and $\J{1}{\Vector{}{\omega}{}}$:
\begin{equation}
    \Q{1}{\Vector{}{\omega}{}}{\Vector{}{z}{}} = H_1(\Vector{}{\omega}{}, \Vector{}{z}{}) + (\J{1}{\Vector{}{\omega}{}}\Vector{}{z}{})^\wedge\J{1}{\Vector{}{\omega}{}} .
    \label{eq:q1}
\end{equation}
Consider the following relations between the left and right Jacobian~\cite{Barfoot2024StateRobotics}:
\begin{align}
    &\Jr{\Vector{}{x}{}} = \Jl{-\Vector{}{x}{}} , &&\Jr{\Vector{}{x}{}} = \AdMsym{\exp(\Vector{}{x}{}^\wedge)^{-1}}\Jl{\Vector{}{x}{}} .
    \label{eq:JlJr}
\end{align}
Recall the following equality which comes from the expansion of the second equation in~\eqref{eq:JlJr}:
\begin{equation}
    \Q{1}{-\Vector{}{\omega}{}}{-\Vector{}{z}{}} = \exp(\Vector{}{\omega}{}^\wedge)^{-1}\Q{1}{\Vector{}{\omega}{}}{\Vector{}{z}{}} - \exp(\Vector{}{\omega}{}^\wedge)^{-1}(\J{1}{\Vector{}{\omega}{}}\Vector{}{z}{})^\wedge\J{1}{\Vector{}{\omega}{}} .
\end{equation}
Then, by substituting~\eqref{eq:q1}, the following equality holds:
\begin{equation}
    \Q{1}{-\Vector{}{\omega}{}}{-\Vector{}{z}{}} = \exp(\Vector{}{\omega}{}^\wedge)^{-1} H_1(\Vector{}{\omega}{}, \Vector{}{z}{}) .
    \label{eq:q1r}
\end{equation}
The $\SEtwo{3}$ right Jacobian matrix can then be written as
\begin{align}
\Jr{\Vector{}{x}{}} &= 
\begin{bmatrix}
    \J{1}{-\Vector{}{\omega}{}} & \zeronm{3}{3} & \zeronm{3}{3}\\
    \exp(\Vector{}{\omega}{}^\wedge)^{-1} H_1(\Vector{}{\omega}{}, \Vector{}{v}{}) & \J{1}{-\Vector{}{\omega}{}} & \zeronm{3}{3}\\
    \exp(\Vector{}{\omega}{}^\wedge)^{-1} H_1(\Vector{}{\omega}{}, \Vector{}{r}{}) & \zeronm{3}{3} & \J{1}{-\Vector{}{\omega}{}}
\end{bmatrix} \\
&= 
\begin{bmatrix}
    \exp(\Vector{}{\omega}{}^\wedge)^{-1}\J{1}{\Vector{}{\omega}{}} & \zeronm{3}{3} & \zeronm{3}{3}\\
    \exp(\Vector{}{\omega}{}^\wedge)^{-1} H_1(\Vector{}{\omega}{}, \Vector{}{v}{}) & \exp(\Vector{}{\omega}{}^\wedge)^{-1}\J{1}{\Vector{}{\omega}{}} & \zeronm{3}{3}\\
    \exp(\Vector{}{\omega}{}^\wedge)^{-1} H_1(\Vector{}{\omega}{}, \Vector{}{r}{}) & \zeronm{3}{3} & \exp(\Vector{}{\omega}{}^\wedge)^{-1}\J{1}{\Vector{}{\omega}{}}
\end{bmatrix} .
\end{align}

\subsubsection{$\G{3}$}
Let $\Vector{}{x}{} = (\Vector{}{\omega}{}, \Vector{}{v}{}, \Vector{}{r}{}, \alpha) \in \Rn{10}$ such that $\Vector{}{x}{}^\wedge \in \g{3}$.
The exponential map of $\G{3}$ is defined as in~\eqref{eq:expG3}, and
the $\G{3}$ left Jacobian can be represented in closed-form as in~\eqref{eq:JlG3}:
\begin{equation}
\Jl{\Vector{}{x}{}} = 
\begin{bmatrix}
    \J{1}{\Vector{}{\omega}{}} & \zeronm{3}{3} & \zeronm{3}{3} & \zeronm{3}{1}\\
    \Q{1}{\Vector{}{\omega}{}}{\Vector{}{v}{}} & \J{1}{\Vector{}{\omega}{}} & \zeronm{3}{3} & \zeronm{3}{1}\\
    \mathbf{\Omega}(\Vector{}{\omega}{}, \Vector{}{v}{}, \Vector{}{r}{}, \alpha) & -\alpha(\J{1}{\Vector{}{\omega}{}} - \J{2}{\Vector{}{\omega}{}}) & \J{1}{\Vector{}{\omega}{}} & \J{2}{\Vector{}{\omega}{}}\Vector{}{v}{}\\
    \zeronm{1}{3} & \zeronm{1}{3} & \zeronm{1}{3} & 1
\end{bmatrix} ,
\end{equation}
where $\mathbf{\Omega}(\Vector{}{\omega}{}, \Vector{}{v}{}, \Vector{}{r}{}, \alpha) = \Q{1}{\Vector{}{\omega}{}}{\Vector{}{r}{}}-\alpha\Q{2}{\Vector{}{\omega}{}}{\Vector{}{v}{}}$.
We already derived an expression for $\mathbf{Q}_1$ in~\eqref{eq:q1}, but in order to find the corresponding alternative closed-form of $\mathbf{\Omega}(\Vector{}{\omega}{}, \Vector{}{v}{}, \Vector{}{r}{}, \alpha)$ we need to express $\Q{2}{\Vector{}{\omega}{}}{\Vector{}{v}{}}$ in terms of $\J{1}{\Vector{}{\omega}{}}, \J{2}{\Vector{}{\omega}{}}$, $H_1(\Vector{}{\omega}{}, \Vector{}{z}{})$ and $H_2(\Vector{}{\omega}{}, \Vector{}{z}{}) \coloneqq \frac{\partial}{\partial \Vector{}{\omega}{}}(\J{2}{\Vector{}{\omega}{}}\Vector{}{z}{})$.
The closed-form expression of $H_2(\Vector{}{\omega}{}, \Vector{}{z}{})$ is
\begin{equation}
    H_2(\Vector{}{\omega}{}, \Vector{}{z}{}) \coloneqq \frac{\partial}{\partial \Vector{}{\omega}{}}(\J{2}{\Vector{}{\omega}{}}\Vector{}{z}{}) = - \kappa_5 \Vector{}{\omega}{}^\wedge\Vector{}{z}{}\Vector{}{\omega}{}^\top - \kappa_2 \Vector{}{z}{}^\wedge - \kappa_6 \Vector{}{\omega}{}^\wedge\Vector{}{\omega}{}^\wedge\Vector{}{z}{}\Vector{}{\omega}{}^\top + \kappa_3 H_0(\Vector{}{\omega}{}, \Vector{}{z}{}) ,
\end{equation}
with
\begin{align}
    &\kappa_3 = \frac{1 - 2\kappa_1}{2 \norm{\Vector{}{\omega}{}}^2}, &\kappa_6 = \frac{\kappa_2 - 2\kappa_4}{\norm{\Vector{}{\omega}{}}^2},
    \label{eq:k36}
\end{align}
and the other terms previously defined in~\eqref{eq:h0} and~\eqref{eq:k01245}.
Then we can derive the following expression for $\mathbf{Q}_2$:
\begin{equation}
\begin{split}
    &\Q{2}{\Vector{}{\omega}{}}{\Vector{}{z}{}} = \Q{1}{\Vector{}{\omega}{}}{\Vector{}{z}{}} - H_2(\Vector{}{\omega}{}, \Vector{}{z}{}) - (\J{2}{\Vector{}{\omega}{}}\Vector{}{z}{})^\wedge\J{1}{\Vector{}{\omega}{}} .
\end{split}
\end{equation}
The final, expanded form of $\mathbf{\Omega}(\Vector{}{\omega}{}, \Vector{}{v}{}, \Vector{}{r}{}, \alpha)$ is
\begin{equation}
    \mathbf{\Omega}(\Vector{}{\omega}{}, \Vector{}{v}{}, \Vector{}{r}{}, \alpha) = H_1(\Vector{}{\omega}{}, \Vector{}{r}{}) + (\J{1}{\Vector{}{\omega}{}}\Vector{}{r}{})^\wedge\J{1}{\Vector{}{\omega}{}} - \alpha (H_1(\Vector{}{\omega}{}, \Vector{}{v}{}) + (\J{1}{\Vector{}{\omega}{}}\Vector{}{v}{})^\wedge\J{1}{\Vector{}{\omega}{}} - H_2(\Vector{}{\omega}{}, \Vector{}{v}{}) - (\J{2}{\Vector{}{\omega}{}}\Vector{}{v}{})^\wedge\J{1}{\Vector{}{\omega}{}}) .
\end{equation}
By applying the relation~\eqref{eq:JlJr}, one can see that the equality~\eqref{eq:q1r} holds also for $\mathbf{Q}_2$:
\begin{equation}
    \Q{2}{-\Vector{}{\omega}{}}{-\Vector{}{z}{}} = \exp(\Vector{}{\omega}{}^\wedge)^{-1} H_2(\Vector{}{\omega}{}, \Vector{}{z}{}) .
    \label{eq:q2r}
\end{equation}
Thus we have that
\begin{equation}
    \mathbf{\Omega}(-\Vector{}{\omega}{}, -\Vector{}{v}{}, -\Vector{}{r}{}, -\alpha) = \exp(\Vector{}{\omega}{}^\wedge)^{-1} (H_1(\Vector{}{\omega}{}, \Vector{}{r}{}) + \alpha H_2(\Vector{}{\omega}{}, \Vector{}{v}{})) .
\end{equation}
Finally, we can express the $\G{3}$ right Jacobian as
\begin{align}
&\Jr{\Vector{}{x}{}} = 
\begin{bmatrix}
    \J{1}{-\Vector{}{\omega}{}} & \zeronm{3}{3} & \zeronm{3}{3} & \zeronm{3}{1}\\
    \exp(\Vector{}{\omega}{}^\wedge)^{-1} H_1(\Vector{}{\omega}{}, \Vector{}{v}{}) & \J{1}{-\Vector{}{\omega}{}} & \zeronm{3}{3} & \zeronm{3}{1}\\
    \exp(\Vector{}{\omega}{}^\wedge)^{-1} (H_1(\Vector{}{\omega}{}, \Vector{}{r}{}) + \alpha H_2(\Vector{}{\omega}{}, \Vector{}{v}{})) & \alpha(\J{1}{-\Vector{}{\omega}{}} - \J{2}{-\Vector{}{\omega}{}}) & \J{1}{-\Vector{}{\omega}{}} & -\J{2}{-\Vector{}{\omega}{}}\Vector{}{v}{} \\
    \zeronm{1}{3} & \zeronm{1}{3} & \zeronm{1}{3} & 1
\end{bmatrix} \\
&= 
\begin{bmatrix}
    \exp(\Vector{}{\omega}{}^\wedge)^{-1}\J{1}{\Vector{}{\omega}{}} & \zeronm{3}{3} & \zeronm{3}{3} & \zeronm{3}{1}\\
    \exp(\Vector{}{\omega}{}^\wedge)^{-1} H_1(\Vector{}{\omega}{}, \Vector{}{v}{}) & \exp(\Vector{}{\omega}{}^\wedge)^{-1}\J{1}{\Vector{}{\omega}{}} & \zeronm{3}{3} & \zeronm{3}{1}\\
    \exp(\Vector{}{\omega}{}^\wedge)^{-1} (H_1(\Vector{}{\omega}{}, \Vector{}{r}{}) + \alpha H_2(\Vector{}{\omega}{}, \Vector{}{v}{})) & \alpha \exp(\Vector{}{\omega}{}^\wedge)^{-1} \J{2}{\Vector{}{\omega}{}} & \exp(\Vector{}{\omega}{}^\wedge)^{-1}\J{1}{\Vector{}{\omega}{}} & -\exp(\Vector{}{\omega}{}^\wedge)^{-1} (\J{1}{\Vector{}{\omega}{}} - \J{2}{\Vector{}{\omega}{}}) \Vector{}{v}{} \\
    \zeronm{1}{3} & \zeronm{1}{3} & \zeronm{1}{3} & 1
\end{bmatrix} .
\end{align}
Note that $\J{1}{-\Vector{}{\omega}{}} = \exp(\Vector{}{\omega}{}^\wedge)^{-1} \J{1}{\Vector{}{\omega}{}}$ and $\J{2}{-\Vector{}{\omega}{}} = \exp(\Vector{}{\omega}{}^\wedge)^{-1} (\J{1}{\Vector{}{\omega}{}} - \J{2}{\Vector{}{\omega}{}})$.